%% file: main.tex
\documentclass[review, 1p]{elsarticle}

\usepackage{mypreamble}
\usepackage{multirow}
\usepackage{setspace}
\usepackage{tabularx}
\usepackage{booktabs}
\usepackage{color}
\usepackage{soul}
\usepackage{comment}
\usepackage{float}
\input{myglossary.tex}

\begin{document}

    \begin{frontmatter}

\title{Subject-Specific Predictive Musculoskeletal Simulations of Lower-Limb Exoskeleton Assistance: Metabolic and Biomechanical Effects of Joint Assistance Strategies}

    
        \author[label1,label2]{Neethan Ratnakumar}

        \author[label1]{Mariya Tohfafarosh}
        
        \author[label1]{Xianlian Zhou\corref{mycorrespondingauthor}}
        \ead{alexzhou@njit.edu}

        \address[label1]{Department of Biomedical Engineering, New Jersey Institute of Technology, Newark, New Jersey, 07102, USA}
    
        \address[label2]{Department of Mechanical Engineering, University of Jaffna, Sri Lanka}
    
        \cortext[mycorrespondingauthor]{Corresponding author}
    
        \begin{abstract}
            Lower-limb exoskeletons have made considerable progress in reducing energy expenditure during walking. However, designing optimal assistance strategies remains challenging, particularly given inter-individual variability in anthropometry and biomechanics. This study explores energy-optimal lower-limb joint-assistance strategies using predictive simulations with musculoskeletal models. Subject-specific models of six able-bodied subjects, with BMI-based muscle strength scaling, were used in predictive simulations to generate gait at self-selected walking speeds. Ideal actuators were incorporated to simulate various combinations of joint assistance at peak levels of $25~\mathrm{Nm}$ and $50~\mathrm{Nm}$ to examine the effects of assistance on gait and metabolic savings. The effects of each assistance configuration were assessed through cost of transport (\acs{COT}), joint kinematics, muscle activations, assistive torques, and joint-level power metrics to characterize the biomechanical and energetic impacts of different assistance strategies. At $50~\mathrm{Nm}$, combined H+K+A (hip–knee–ankle) assistance resulted in the greatest mean COT reduction of 48.50 $\pm$ 4.55\%, with individual reductions ranging from 42.22\% to 54.65\%  across the subjects. Among single joint conditions, assisting the hip was most effective, reducing COT by 31.77 $\pm$ 5.21\%; the knee and ankle produced smaller, comparable reductions (19.17\% and 17.02\%). H+A (hip-ankle) assistance (44.60 $\pm$ 7.08\%) emerged as the most effective two-joint assistive configuration. Increasing the torque bound increased positive assistive power primarily at the hip and ankle, while knee assistance showed little sensitivity and delivered positive power near pre-swing. These results support H+A assistance as an efficient two-actuator target, while identifying the knee's atypical pre-swing power strategy as a candidate for targeted experimental validation.
        \end{abstract}
        
    
    \end{frontmatter}

    \input{myacronyms.tex}

    \doublespacing


    \acresetall

    \section{Introduction}\label{sec:introduction}

        Lower-limb exoskeletons have transformed gait assistive technology, demonstrating the ability to reduce metabolic energy expenditure~\cite{zhou2024review, Sawicki2020,luo2024experiment} and muscle activations~\cite{Xu2023, Wang2025, Durandau2022} around assisted joints while also being able to increase walking speed~\cite{lakmazaheri2024optimizing, song2021optimizing}. These advancements underscore the potential of exoskeletons to significantly improve human mobility and quality of life. However, despite these advancements, designing individualized optimal assistance remains challenging. Wearers exhibit varied physiological, biomechanical, and neurological responses to assistance, influenced by factors like muscle strength,  \ac{BMI}, body segment lengths, sex, and other individual differences. Tailored assistance strategies are essential for addressing this variability and optimizing exoskeleton performance.
        
        Early exoskeleton studies adjusted torque magnitude and timing through expert-guided or heuristic tuning~\cite{young2017influence, quinlivan2017assistance, galle2017reducing, ding2016effect, lee2017effects, kantharaju2023framework,lopes2022deep,echeveste2025emg}. More recently, Human in the Loop (\acs{HIL}) optimization has used physiological or biomechanical feedback to search the assistance-parameter space more systematically~\cite{zhang2017human,franks2021comparing,monteiro2024human,kim2017human,kim2022reducing,ding2018human}. Both optimized parameters and user preferences vary across individuals~\cite{lee2023user,ingraham2022role}, motivating approaches that can evaluate individualized assistance without extensive experimental tuning.
    	
        Musculoskeletal simulations have emerged as a complementary approach for developing exoskeleton control strategies. Recent reviews highlight the growing role of simulations in assistive device development, emphasizing their value in exploring control strategies, optimizing mechanical designs, and reducing the need for extensive experimental testing~\cite{firouzi2025biomechanical,grabke2019lower}. Advances in computational methods have enabled predictive simulations with complex neuro-musculoskeletal models, offering a platform to test and refine assistive strategies without the need for extensive human trials. Methods based on prescribed kinematics, including computed muscle control (\acs{CMC}), have been used to estimate muscle forces during exoskeleton-assisted movement~\cite{zhou2020design, dembia2017simulating,uchida2016simulating,zhou2020predictive}. However, these methods rely on fixed kinematics derived from pre-recorded motion data without assistance, which limits their ability to predict how movement patterns dynamically change in response to external assistance. Predictive optimal-control approaches address this limitation by optimizing both movement and assistance. In particular, single-shooting methods have been implemented in tools such as \acs{SCONE}~\cite{geijtenbeek2019scone} with reflex-based neuromuscular control strategies~\cite{geyer2010muscle}. These approaches enable predictive simulations to optimize exoskeleton assistance under realistic conditions. For example, SCONE-based predictive simulations have been used to design assistance modes for walking on slopes~\cite{li2020using}, optimize torque profiles for idealized hip exoskeletons to improve walking efficiency and reduce user effort~\cite{ratnakumar2020optimized}, and evaluate the effects of free versus joint moment assistance during gait to inform effective assistance strategies~\cite{jabeen2019assisting}. Although effective, shooting methods can be computationally expensive, as they adjust control parameters iteratively to minimize the cost function.

    	In recent years, direct collocation methods, which discretize the problem into multiple segments to improve computational efficiency, have gained traction. Direct collocation exploits the sparsity of the problem formulation to significantly reduce the computational cost of simulating human movement compared to traditional methods. The growing ecosystem of open-source tools, such as OpenSim Moco, has made direct collocation frameworks more accessible, enabling researchers to generate predictive simulations that are much faster~\cite{dembia2020opensim}. The work presented in~\cite{dembia2020simulating} highlights the differentiability of direct collocation, enabling efficient integration of dynamic constraints. This makes it highly effective for simulating complex human movements with accuracy. Similarly, PredSim, a tool developed by D'Hondt et al.~\cite{PredSim2024}, incorporates algorithmic differentiation and direct collocation to enable robust predictive simulations at higher speed. 
        
    	However, the ability of predictive simulations to accurately replicate individual-level adaptations remains limited. Weiss et al.~\cite{weiss2024pairing} paired subject-specific predictive simulations with experimental data on ankle exoskeleton walking. While they successfully matched the average reductions in metabolic cost (13\%) and ankle kinematics across participants, they observed significant discrepancies at the individual level. For instance, participants who had high experimental cost savings were often predicted to have low savings, and vice versa. Furthermore, simulated kinetics deviated from measured values, especially during exoskeleton-assisted walking. These findings suggest that current models, even when scaled to individual anthropometry, fall short in capturing the nuanced ways people adapt to external torque, highlighting the complexity of subject-specific human-exoskeleton interactions and the need for more accurate, personalized simulation frameworks.
        
        To address this gap, we present, to our knowledge, the first systematic evaluation of the full combinatorial space of single- and multi-joint hip, knee, and ankle exoskeleton assistance, seven configurations at two torque levels, across a cohort of subject-specific musculoskeletal models spanning a broad range of anthropometry, in a predictive musculoskeletal simulation framework. This paper investigates subject-specific predictive gait assistive simulations, focusing on how individual differences in anthropometry can influence both biomechanical and metabolic outcomes, as well as the resulting assistive torque profiles. This exploration is critical to determine whether the generalized findings from generic predictive simulations~\cite{ratnakumar2020optimized, ratnakumar2022prediction} are applicable to individuals with diverse body profiles. We generate predictive simulations for multiple subject-specific lower-limb musculoskeletal models, each featuring unique anthropometric characteristics. Starting from unassisted walking, we simulate eight conditions in total: unassisted (NoExo), and seven assistance modes applying torque bilaterally at the hip, knee, and ankle joints, both individually and in all possible combinations. We then compare the resulting individualized assistance strategies, evaluating key measures such as gait kinematics, metabolic cost, muscle activation, joint torque and power. Finally, we provide a detailed discussion of the results and their implications before concluding.

    \section{Methodology}\label{sec:methodology}

\begin{figure}[htbp]
    \centering
    \includegraphics[width=1.0\linewidth]
    {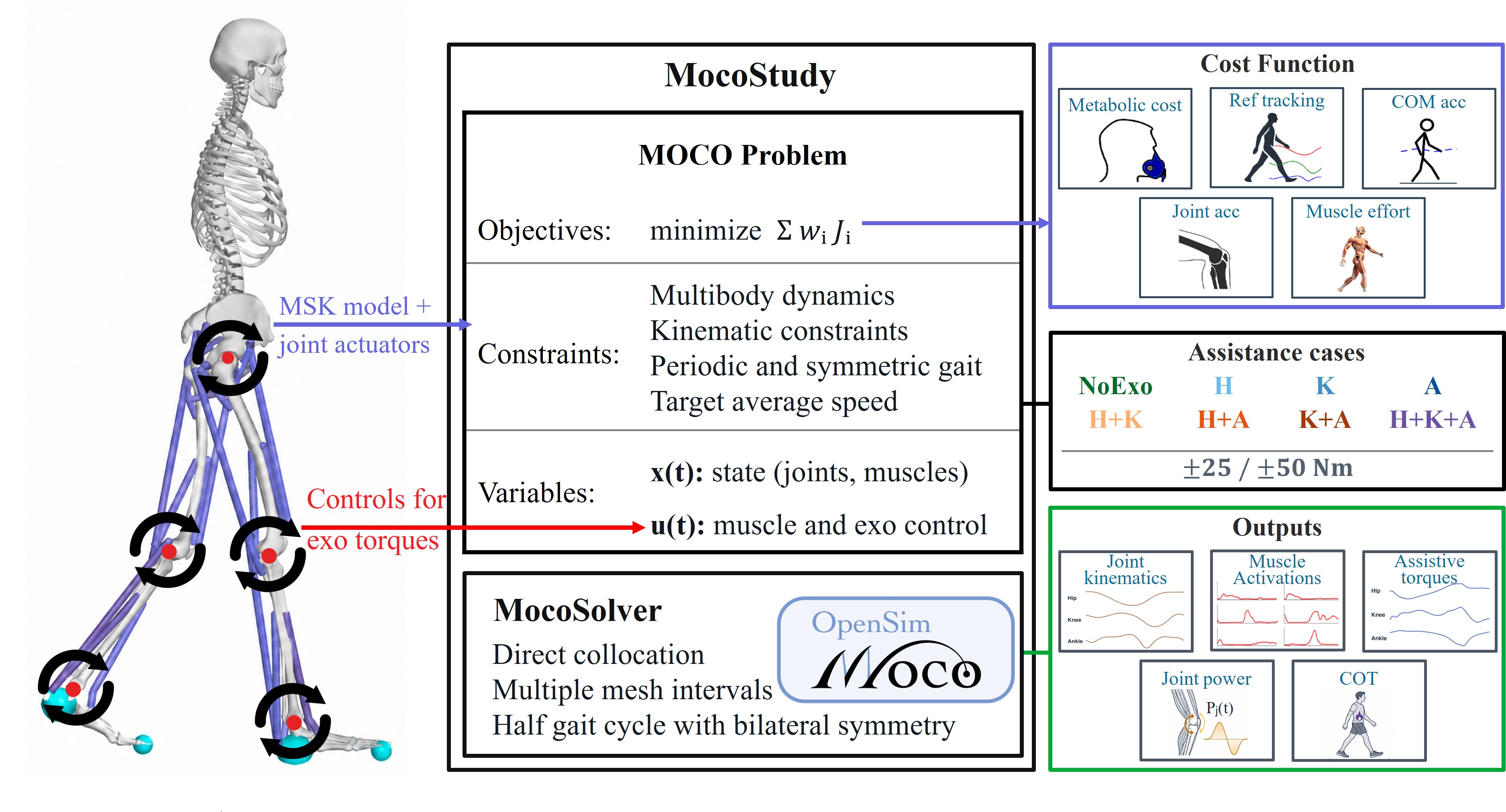}
    \caption{Overview of the OpenSim Moco optimal-control framework used
    to generate the unassisted and exoskeleton-assisted predictive
    walking simulations. Exoskeleton assistance is modeled as ideal torque actuation at the joint coordinates, optimized together with other variables such as muscle control and joint angles.}
    \label{fig:moco_framework}
\end{figure}

    Musculoskeletal optimal control simulations were performed to study various exoskeleton-assisted conditions for energy-optimized walking. Using the OpenSim Moco framework~\cite{dembia2020opensim}, simulations were conducted for unassisted walking (NoExo) and seven single and multi-joint bidirectional torque assistance conditions. An overview of the OpenSim Moco optimization framework used for our simulations is shown in Figure~\ref{fig:moco_framework}. All simulations were formulated as predictive optimal control problems that incorporate a kinematic tracking term, with reference trajectories taken from the corresponding overground gait data in an open-source lower-limb dataset~\cite{camargo2021comprehensive}. 

    \subsection{Musculoskeletal model}\label{Generic-model}
\begin{figure}[htbp]
        \centering      
        \includegraphics[width=1.0\textwidth]{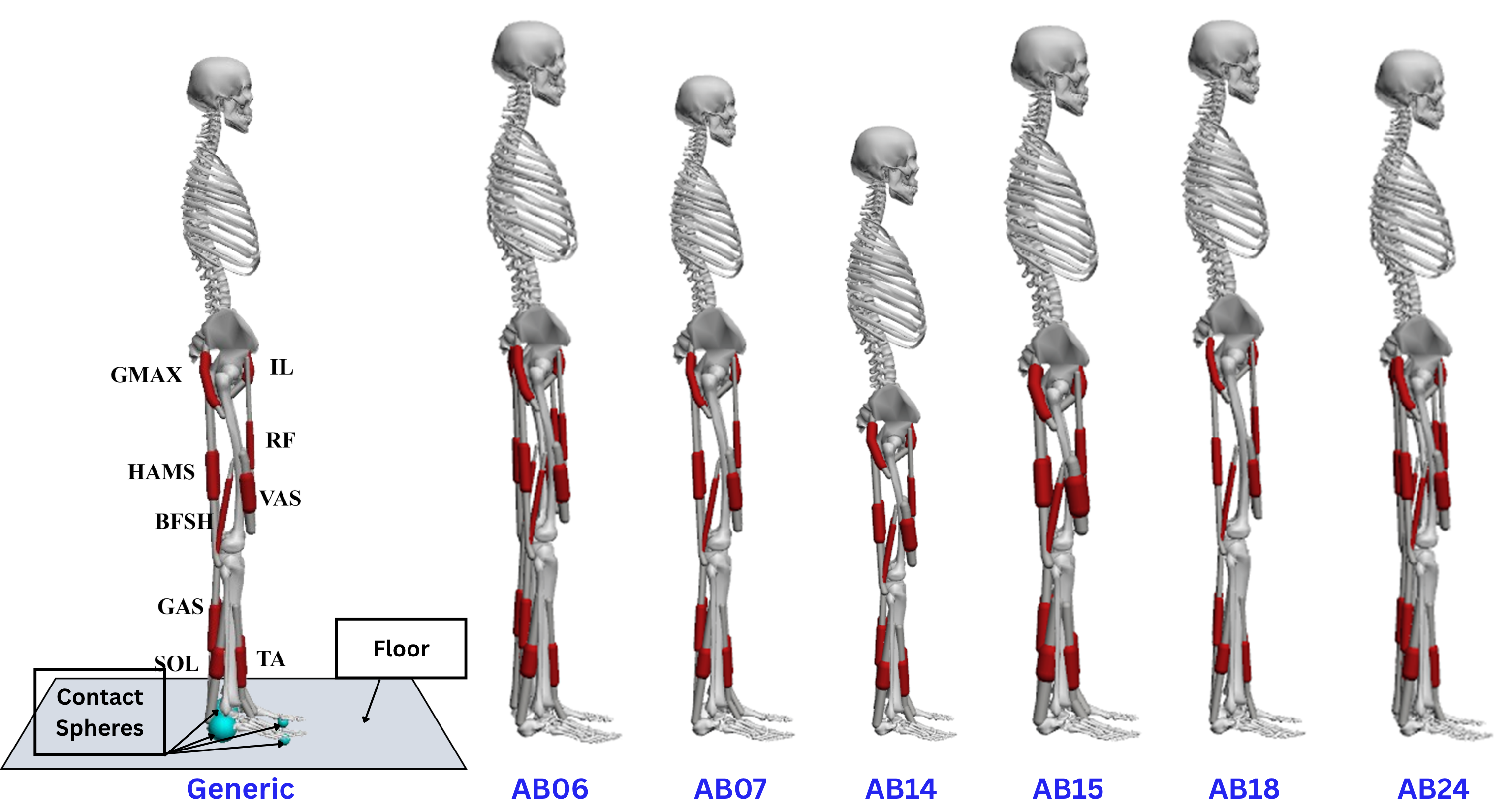}
        \caption{The generic and scaled musculoskeletal models. The cross-sectional size of each muscle is scaled proportionally to its maximum isometric force, with floor contact represented by contact spheres.}
        \label{fig:generic}
    \end{figure}

    A 2D generic musculoskeletal model that is bilaterally symmetric was obtained from the OpenSim~\cite{delp2007opensim} release and selected for the simulations. 
    The model consists of 10 degrees of freedom (\acs{DOF}), comprising the lumbar joint, bilateral hips, knees, ankles, and three global DOFs (pelvis tilt, and horizontal and vertical pelvis translation). This model represents an adult with a body mass of 62 kg and an approximate height of 1.70 m, corresponding to a mid-range body mass index (BMI = 21.45). The model includes 18 musculotendon units, with 9 muscles per limb. These include the gluteus maximus (GMAX), iliopsoas (IL), hamstrings (HAMS), rectus femoris (RF), vasti (VAS), biceps femoris short head (BFSH), gastrocnemius (GAS), tibialis anterior (TA), and soleus (SOL). The lumbar joint features a pin joint actuated by a coordinate actuator capable of delivering peak torques up to $\pm 150$ Nm, compensating for the absence of muscles spanning the lumbar joint. The front and side views of this generic musculoskeletal model are shown in Figure~\ref{fig:generic}.

    The DeGrooteFregly muscle model~\cite{degroote2016evaluation}, which is known for its smooth and differentiable formulation, was utilized to support gradient-based optimization methods. Muscle dynamics were calculated by assuming rigid tendons, with tendon compliance ignored to reduce computational complexity. The maximum isometric forces, optimal fiber lengths, and tendon slack lengths for these muscles are listed in the appendix (Table \ref{tab:generic_model_muscle_properties}). Activation and deactivation time constants were set to 15 ms and 60 ms, respectively. A baseline minimum activation level of 0.01 was imposed on all muscles to prevent numerical instability during simulations. The maximum contraction velocity for all muscles was standardized at 10 optimal fiber lengths per second.
    Similarly, smooth contact interactions between the ground and contact spheres positioned on the heels and forefeet were modeled using a modified version of the Hunt-Crossley contact model~\cite{hunt1975coefficient}. The implementation utilized OpenSim SmoothSphereHalfSpaceForce components, in which spherical geometries representing the foot contact points interacted with the ground surface represented as a ContactHalfSpace. The normal contact stiffness was set to \( 3 \times 10^6 \) N/m\(^2\). The velocity-dependent damping was modeled using a dissipation coefficient of $2~\mathrm{s/m}$. Tangential interactions and sliding behavior were characterized by static and dynamic friction coefficients of 0.8 and a viscous friction coefficient of 0.5. The smoothness of the transition between the Hertzian and Hunt-Crossley forces was controlled through the \texttt{hertz\_smoothing} and \texttt{hunt\_crossley\_smoothing} parameters, set to 300 and 50, respectively.

    \subsection{Subject-specific models}\label{subject-specifc-model}
    Subject-specific models were created by anthropometrically scaling the generic model to match the height, body weight, and segment lengths of six subjects (three males and three females) selected from the open-source database reported in~\cite{camargo2021comprehensive}. The six subjects had an average height of $1.71 \pm 0.11$ m and an average body mass of $69.58 \pm 15.23$ kg. They were selected to represent a broad range of anthropometric characteristics among the 22 subjects in the database, including extremes in height, body mass, and body mass index (BMI). Specifically, AB14 was the shortest subject; AB06 and AB18 were two of the three tallest subjects (height: 1.80 m), with AB18 also having the lowest BMI; AB15 had the highest BMI and body mass; AB07 was the shortest and lightest male; and AB24 was the heaviest female. Additional details on subject demographics are provided in Table~\ref{tab:subject_demographics}. The scaling process was performed using the OpenSim Scale Tool to provide subject-specific representations of musculoskeletal geometry.
    
    \input{subject_demo}
        
    In addition to geometric scaling, the maximum isometric muscle forces (\acs{F-max}) of each muscle were scaled based on the subject's height (\acs{H}) and body mass (\acs{M}) using the following equation:
    \begin{equation}
        \acs{F-max} = \acs{F-max-G} \times \frac{\acs{M}/\acs{H}}{\acs{M-G}/\acs{H-G}} \times \frac{\acs{R-MSK}}{\acs{R-MSK-G}}
    \end{equation}
    where superscript $G$ indicates the generic model's corresponding variables. $M/H$ is the normalized mass per unit of height, approximately proportional to the average body cross-sectional area. \acs{R-MSK} and \acs{R-MSK-G} are the muscle-to-mass ratio of the subject and the generic model, respectively. 
    The muscle-to-mass ratio was calculated as $\acs{R-MSK} = 1 - R_{\text{other}} - R_{\text{fat}}$, where $R_{\text{other}}$ represents the proportion of mass not attributed to muscle, fat, or bone and was set to 0.5. The fat mass ratio was estimated using the subject's body mass index (BMI) as $R_{\text{fat}} = \left( -0.09 + 0.0149 \times \text{BMI} - 0.00009 \times \text{BMI}^2 \right)$~\cite{wang2022musculoskeletal}.  
    The computed values of $F_{max}$ for the subject-specific models are listed in Appendix Table \ref{tab:max_isometric_muscle_forces}. 
    
    \subsection{Predictive simulations}\label{predictive-simulations-with-tracking-term}
    
    Optimization of walking with and without assistance was formulated as an optimal control problem within OpenSim Moco~\cite{dembia2020opensim}. The optimization variables include controls and states (i.e., muscle controls or excitations, joint coordinate angles, and assistance torques as coordinate actuation). The constraints include equations of motion, kinematic constraints (such as symmetry and periodicity), state and control upper and lower bounds, boundary constraints, and time limits, etc~\cite{ratnakumar2022prediction}. Subject-specific models utilized the corresponding overground gait data from the open-source dataset~\cite{camargo2021comprehensive} as the reference walking motion trajectory. 
    
    The objective or cost function employed in this study is defined as follows:
    \begin{equation} 
        J_T = \int_{t=0}^{t_f} \left( w_1 J_{\text{MER}} + w_2 J_{\text{COM}} + w_3 J_{\text{Joint}} + w_4 J_{\text{Effort}} + w_5 J_{\text{Track}} \right) dt \label{eq:total_cost} 
    \end{equation}
    
    The cost function (\acs{J-T}) comprised a metabolic cost term (\acs{J-MER}), a muscle effort term (\acs{J-Effort}), a minimum joint acceleration term (\acs{J-Joint}), a minimum center of mass acceleration term (\acs{J-COM}), and a reference movement tracking term (\acs{J-Track}), as shown in Equation~\eqref{eq:total_cost}. Each cost term was weighted by coefficients $w_1$, $w_2$, $w_3$, $w_4$, and $w_5$. The values for the cost term weights were set as follows: $w_1 = 1$, $w_2 = 0.01$, $w_3 = 2.5 \times 10^{-4}$, $w_4 = 10$, and $w_5 = 10$.
    
    The metabolic cost term, \acs{J-MER}, is the same as the COT, which is defined as the total metabolic cost (the sum of individual muscle metabolic rates over the duration $T$ based on the Uchida energetics model~\cite{uchida2016simulating}) normalized by the distance traveled ($D$) and the model’s body mass ($M$):
    \begin{equation}
        J_{MER} = \frac{1}{M\times D} \sum_{i=1}^{N} \dot{E}_i(t)
    \end{equation}
    This term penalizes the total energy cost per meter walked and accounts for individual muscle contributions through muscle-specific parameters such as slow-twitch fiber ratios and specific tension values. 
    
    The whole-body (\acs{COM}) acceleration term was defined as $J_{\text{COM}} = w_x a_x^2 + w_y a_y^2 + w_z a_z^2$, where $a_x$, $a_y$, and $a_z$ are the (\acs{COM}) accelerations in each direction, and $[w_x,\ w_y,\ w_z] = [30.0,\ 2.0,\ 0.0]$ are the corresponding weights. This formulation penalizes excessive acceleration of the body, which helps reduce vertical ground reaction forces during initial contact. 
    The joint acceleration term was defined as $J_{\text{Joint}} = \sum_i \ddot{\theta}_i^2$, where $\ddot{\theta}_i$ is the acceleration of the $i$-th joint angle. This term discourages rapid changes in joint velocity by penalizing acceleration.
    
    The control effort term was defined as $J_{\text{Effort}} = \sum_i k_i x_i^2$, where $x_i$ represents the control signal (e.g., muscle excitation or joint assistive torque), and $k_i$ is a local weighting coefficient. To isolate muscle effort, $k_i$ was set to 1 for muscles and 0 for exoskeleton actuators and the lumbar coordinate actuator. This use of $k_i$ is distinct from the global cost-term weights $w_1$ to $w_5$ used in the overall objective function.
    
    The simulations were performed using multiple mesh resolutions, ranging from 36 to 50 mesh intervals in increments of 2. In OpenSim Moco, the simulation time horizon is divided into a specified number of mesh intervals, within which the continuous state and control trajectories are discretized and approximated using Moco's default transcription method. Increasing the number of mesh intervals provides a finer temporal discretization of the optimal control problem, potentially allowing the state and control trajectories to be represented with greater resolution, but at the cost of increased computational expense. Simulations were performed at each mesh resolution, and the solution yielding the greatest COT reduction was selected for subsequent analysis.
    The musculoskeletal model was assumed to walk with a bilaterally symmetric gait to improve computational efficiency, as this permitted the simulation of only half of a gait cycle. The unassisted gait simulations were followed by simulations incorporating bilateral idealized joint actuators that provided up to  $\pm 50$ Nm of assistive torque in the sagittal plane. Assistive conditions included hip only (H), knee only (K), ankle only (A), and all possible combinations (H+K, H+A, K+A, H+K+A). In addition to the  $\pm 50$ Nm, we also performed the simulations with  $\pm 25$ Nm. These values were selected to span a broad range of assistive torque magnitudes commonly explored in the exoskeleton literature; low~\cite{seo2016fully, yu2020quasi, lerner2018untethered, gomez2021experimental}, moderate~\cite{ishmael2022powered, slade2022personalizing}, and high~\cite{bajpai2023design, kang2023design} assistive torque levels.

    \subsection{Power analysis}\label{power-analysis}
    A joint power analysis was performed to evaluate the overall mechanical demands and net energy transfer characteristics across the joint. The \ac{MPP} and \ac{MNP} were calculated to evaluate the phases of net energy generation and net energy absorption, respectively, using the following equations~\cite{lim2023parametric}:
    \begin{equation}
        \text{MPP} = 1/n \sum_{i=1}^{n_p} \tau_i \dot{q}_i, \quad (\text{for } \tau_i \dot{q}_i > 0)
    \end{equation}
    
    \begin{equation}
        \text{MNP} = 1/n \sum_{i=1}^{n_n} \tau_i \dot{q}_i \quad (\text{for } \tau_i \dot{q}_i < 0)
    \end{equation}
    
    \noindent where $\tau_i$ is the joint torque and $\dot{q}_i$ is the joint angular velocity at the $i^{\text{th}}$ time step, such that $\tau_i \dot{q}_i$ represents the instantaneous joint power. The terms $n_p$ and $n_n$ denote the number of samples corresponding to positive and negative power, respectively, and $n$ is the total number of samples. 
    In addition, the Positive Power Ratio (\acs{PPR}) was computed to quantify the relative contribution of positive work, defined as:
    \begin{equation}
        \text{PPR} = \frac{\sum_{i} \left( \tau_i \dot{q}_i \;\; \text{for } \tau_i \dot{q}_i > 0 \right)}{\sum_{i} \left| \tau_i \dot{q}_i \right|}
    \label{PPR_Equation}
    \end{equation}

\subsection{Statistical analysis}
\label{sec:statistical-analysis}

COT reductions at $50~\mathrm{Nm}$ were evaluated using three
nonparametric tests. Exact Friedman tests were used to compare the
paired single-joint configurations and the paired two-joint
configurations. An exact paired Wilcoxon signed-rank test compared
H+A with H+K+A. Holm correction was applied across the three tests,
with statistical significance set at $\alpha=0.05$. Full statistical
results are reported in Appendix
Table~\ref{tab:cot_minimal_statistics}.
    
    \section{Results}\label{sec:results}

\begin{figure}[htb!]
        \centering
        \includegraphics[width=0.98\textwidth]{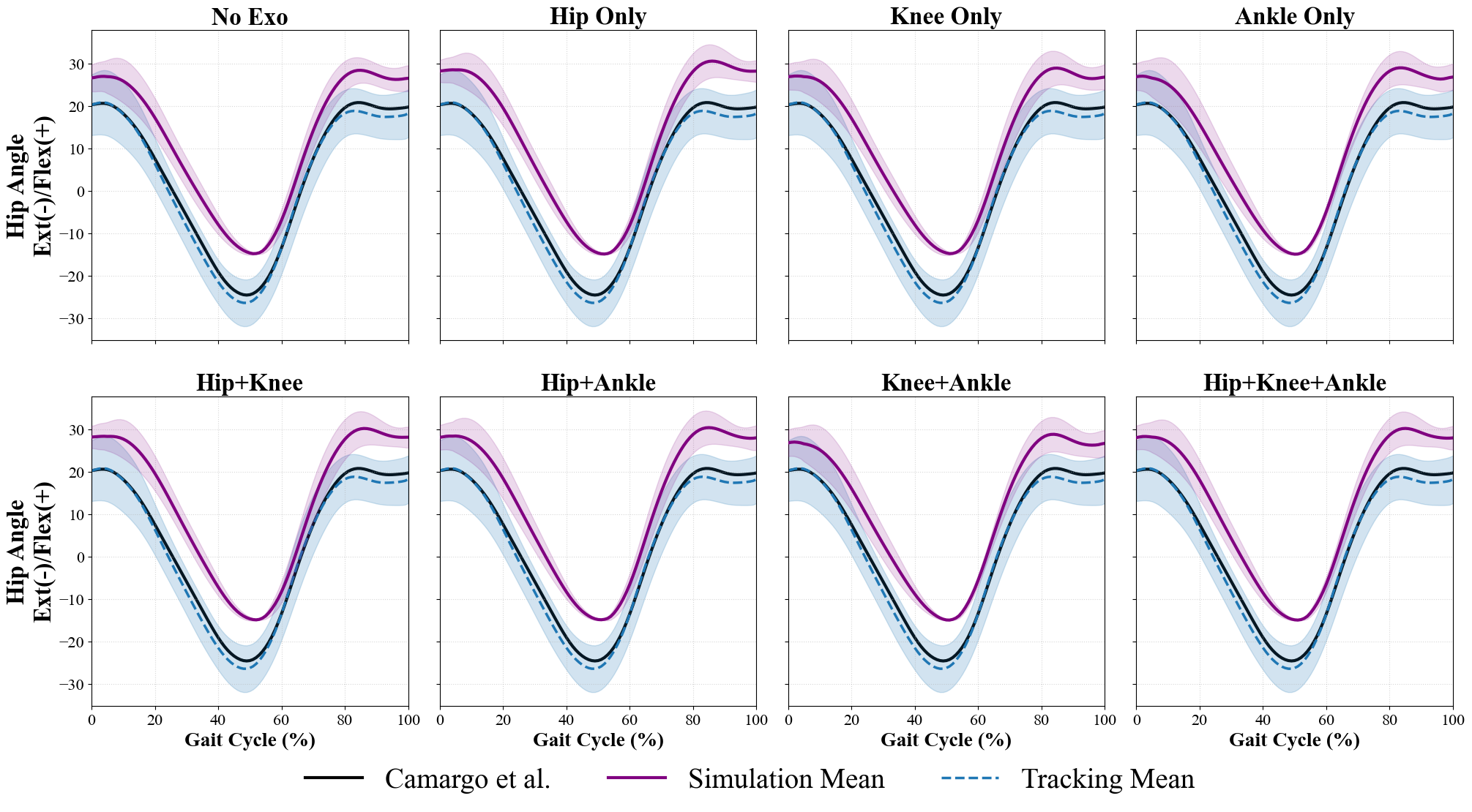}
        \caption{Hip flexion-extension angles from predictive tracking simulations across seven exoskeleton assistance conditions ($50~\mathrm{Nm}$) and the unassisted baseline (NoExo). Each subplot shows the hip angle profile obtained by solving the optimal control problem for the specific condition. The solid purple line and shaded area represent the mean $\pm$ standard deviation across subject-specific simulations. For each subject-specific model, one representative unassisted gait cycle from that subject was used as the tracking reference. The dashed tracking mean data are averaged from the six subjects, and the experimental angle data from all subjects in Camargo et al.~\cite{camargo2021comprehensive} is shown in black.
    }
        \label{fig:hip_angle}
    \end{figure}

\begin{figure}[htb!]
        \centering
        \includegraphics[width=0.98\textwidth]{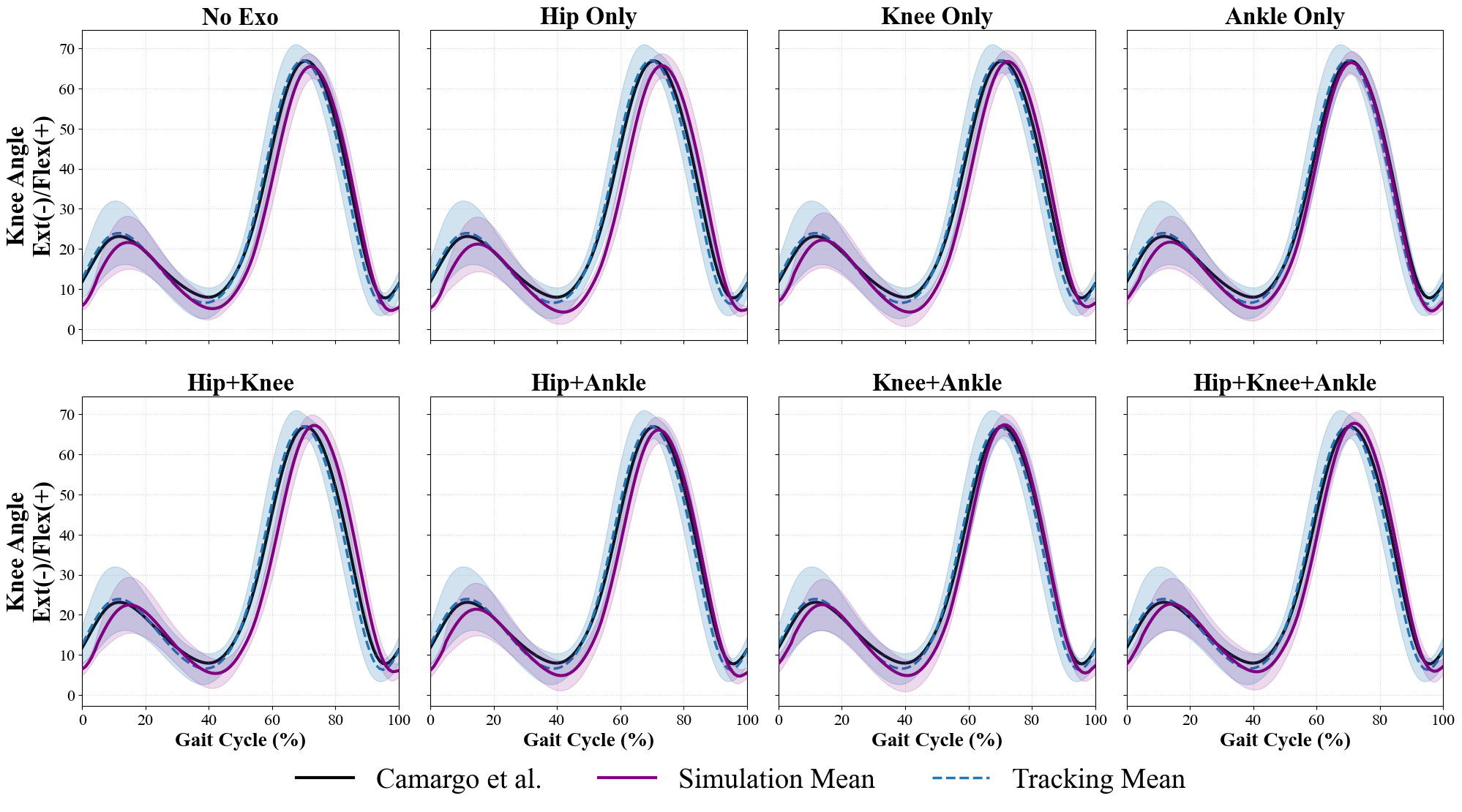}
        \caption{Knee flexion-extension angles resulting from predictive tracking simulations across seven exoskeleton assistance conditions ($50~\mathrm{Nm}$) and the unassisted baseline (NoExo). Each subplot shows the knee angle profile obtained by solving the optimal control problem for the specific condition. The solid purple line and shaded area represent the mean $\pm$ standard deviation across subject-specific simulations. For each subject-specific model, one representative unassisted gait cycle from that subject was used as the tracking reference.  The dashed tracking mean data are averaged from the six subjects, and the experimental angle data from all subjects in Camargo et al.~\cite{camargo2021comprehensive} is shown in black.}
        \label{fig:knee_angle}
    \end{figure}

\begin{figure}[htb!]
        \centering
        \includegraphics[width=0.98\textwidth]{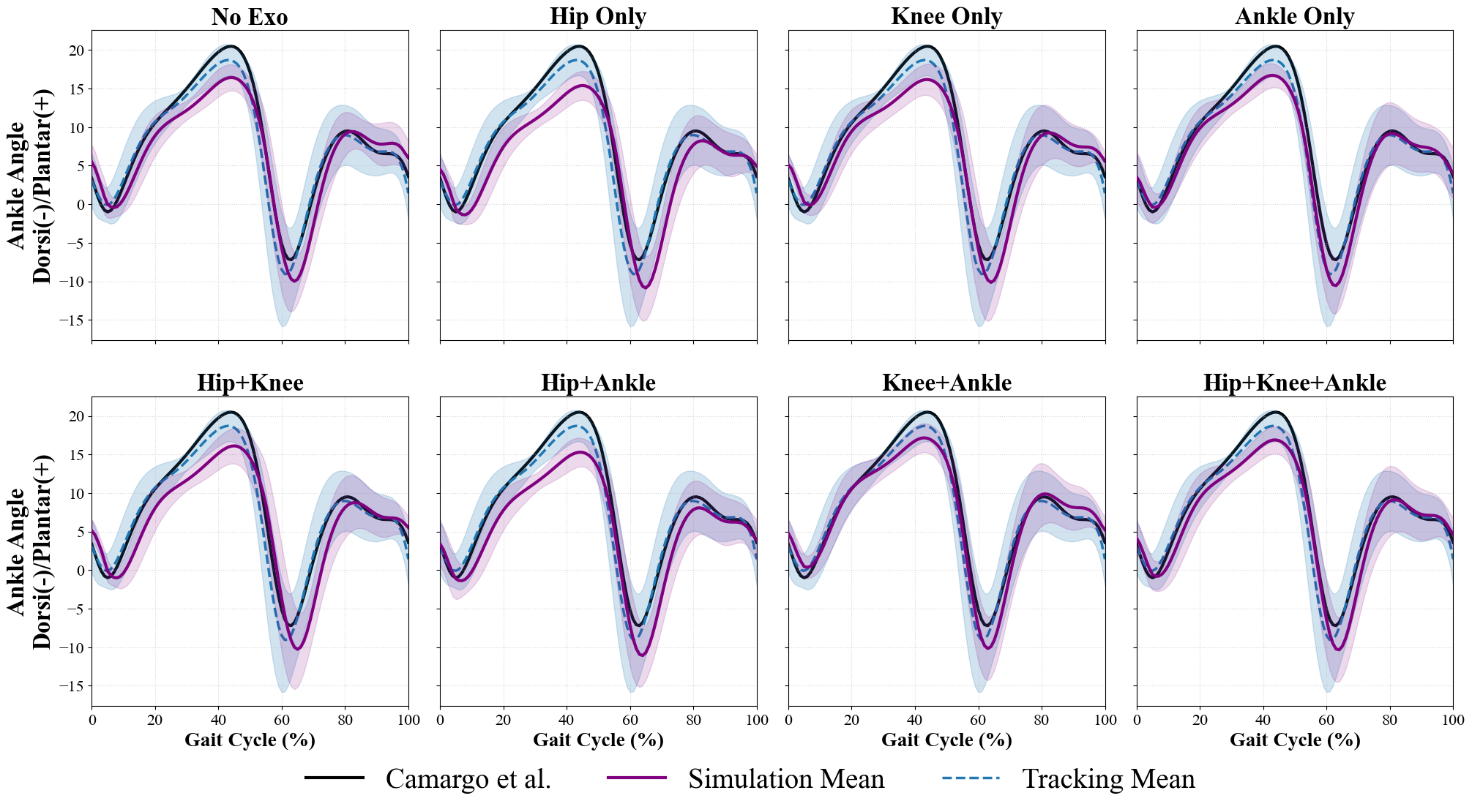}
        \caption{Ankle dorsiflexion-plantarflexion angles resulting from predictive tracking simulations across seven exoskeleton assistance conditions ($50~\mathrm{Nm}$) and the unassisted baseline (NoExo). Each subplot shows the ankle angle profile obtained by solving the optimal control problem for the specific condition. The solid purple line and shaded area represent the mean $\pm$ standard deviation across subject-specific simulations. For each subject-specific model, one representative unassisted gait cycle from that subject was used as the tracking reference.  The dashed tracking mean data are averaged from the six subjects, and the experimental angle data from all subjects in Camargo et al.~\cite{camargo2021comprehensive} is shown in black.}
        \label{fig:ankle_angle}
    \end{figure}

\begin{figure}[htb!]
        \centering
        \includegraphics[width=0.98\textwidth]{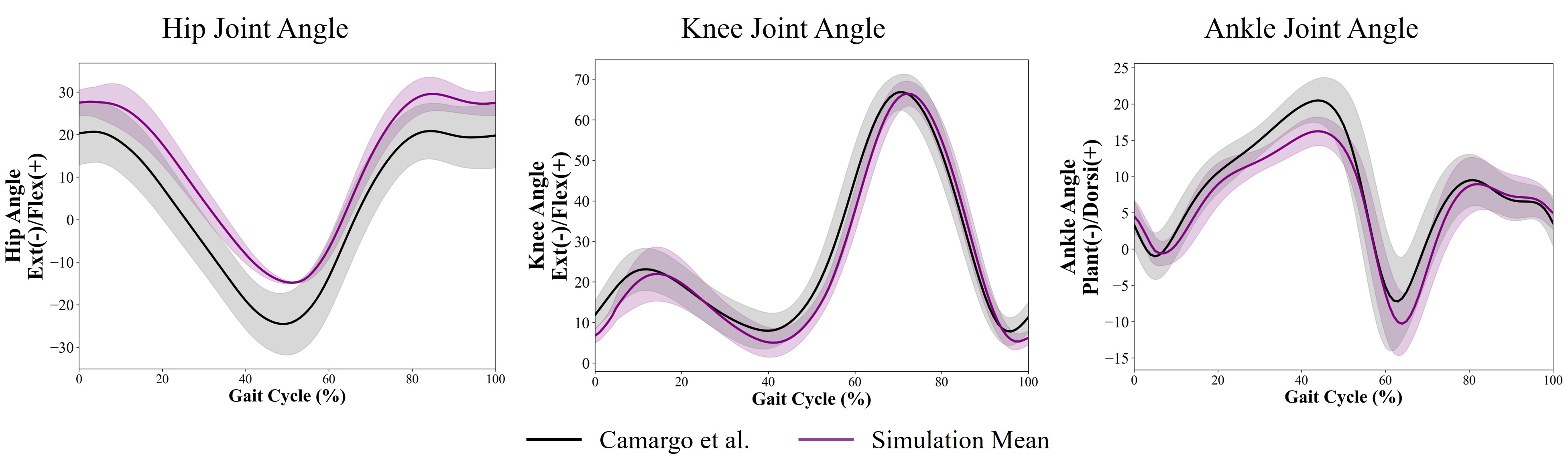}
        \caption{Overall simulated hip, knee, and ankle joint angles averaged across all subject-specific models at $50~\mathrm{Nm}$ assistance condition. 
    The purple solid colored line represents the mean joint angle trajectory, and the shaded area denotes the standard deviation. 
    Experimental data from Camargo et al.~\cite{camargo2021comprehensive} is shown in black with a gray shaded region representing $\pm$1 SD.}
        \label{fig:angles}
    \end{figure}

    Results from predictive simulations were summarized for both the unassisted baseline (NoExo) and all seven exoskeleton-assistance configurations (under $50~\mathrm{Nm}$) across the subject-specific models. The resulting mean joint kinematics from the tracking simulations for the hip, knee, and ankle joints are presented in Figures \ref{fig:hip_angle} through \ref{fig:ankle_angle}, respectively, while the overall aggregated simulation data across all conditions are presented in Figure \ref{fig:angles}. All simulated trajectories were evaluated against the experimental tracking data mean (averaged across the six subjects) and the data from all subjects in Camargo et al.~\cite{camargo2021comprehensive}. 
    In Figure \ref{fig:hip_angle}, there is a clear offset between the simulated mean hip angle and the experimental tracking mean. The maximum extension (~30°) appears in both the Camargo et al.~\cite{camargo2021comprehensive} reference and the six-subject tracking mean, consistent with a hip-flexion coordinate offset in the Camargo inverse-kinematics solution rather than a true difference in hip range of motion. Because the cost function combines kinematic tracking with effort minimization, the optimizer departed from this offset reference and converged toward extension angles (~15°) consistent with the unassisted baseline and the maximum hip extension angle typically reported in the literature~\cite{riglet20243d,fukuchi2018public}. For example, Fukuchi et al.~\cite{fukuchi2018public} reported maximum hip extension angle below or near $10\degree$ for overground walking at three different speeds (slow, normal, and fast). Our simulation results automatically corrected this discrepancy and converged to more realistic hip angle ranges. For the knee angles in Figure \ref{fig:knee_angle}, the predicted data traces the experimental data well, with a very small lateral time offset. The maximum flexion angle reaches around 70 degrees for all conditions. For ankle angles in Figure \ref{fig:ankle_angle}, the predicted data also trace the tracking experimental data well. However, during peak plantarflexion, the predicted data have a smaller angle relative to the tracking data. 
    
    \begin{figure}[htb!]
        \centering
        \includegraphics[width=0.98\textwidth]{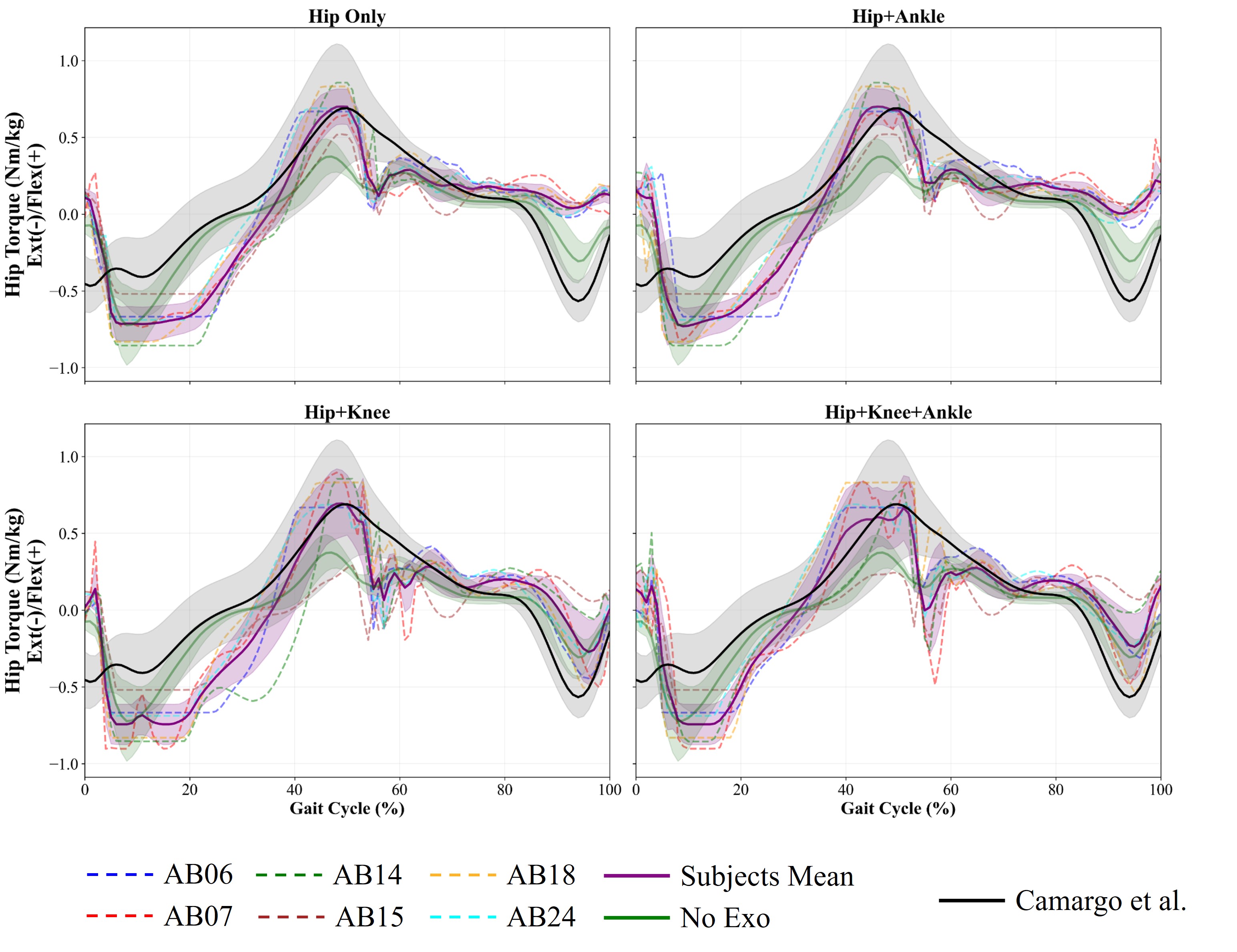}
        \caption{Predicted hip assistive torque ($50~\mathrm{Nm}$) profiles across the gait cycle for the four exoskeleton configurations (Hip Only, H+A, H+K, and H+K+A) that incorporate the hip assistance. Dashed colored lines indicate individual subject data, while the purple line (with shaded $\pm$SD) represents the subject-specific model group average. The black line shows the experimental biological hip joint torque from Camargo et al.~\cite{camargo2021comprehensive}, and the green line (with shaded $\pm$SD) depicts the mean of the NoExo condition for the subject-specific models.}
        \label{fig:hip_torques}
    \end{figure}
    
    \begin{figure}[htb!]
        \centering
        \includegraphics[width=0.98\textwidth]{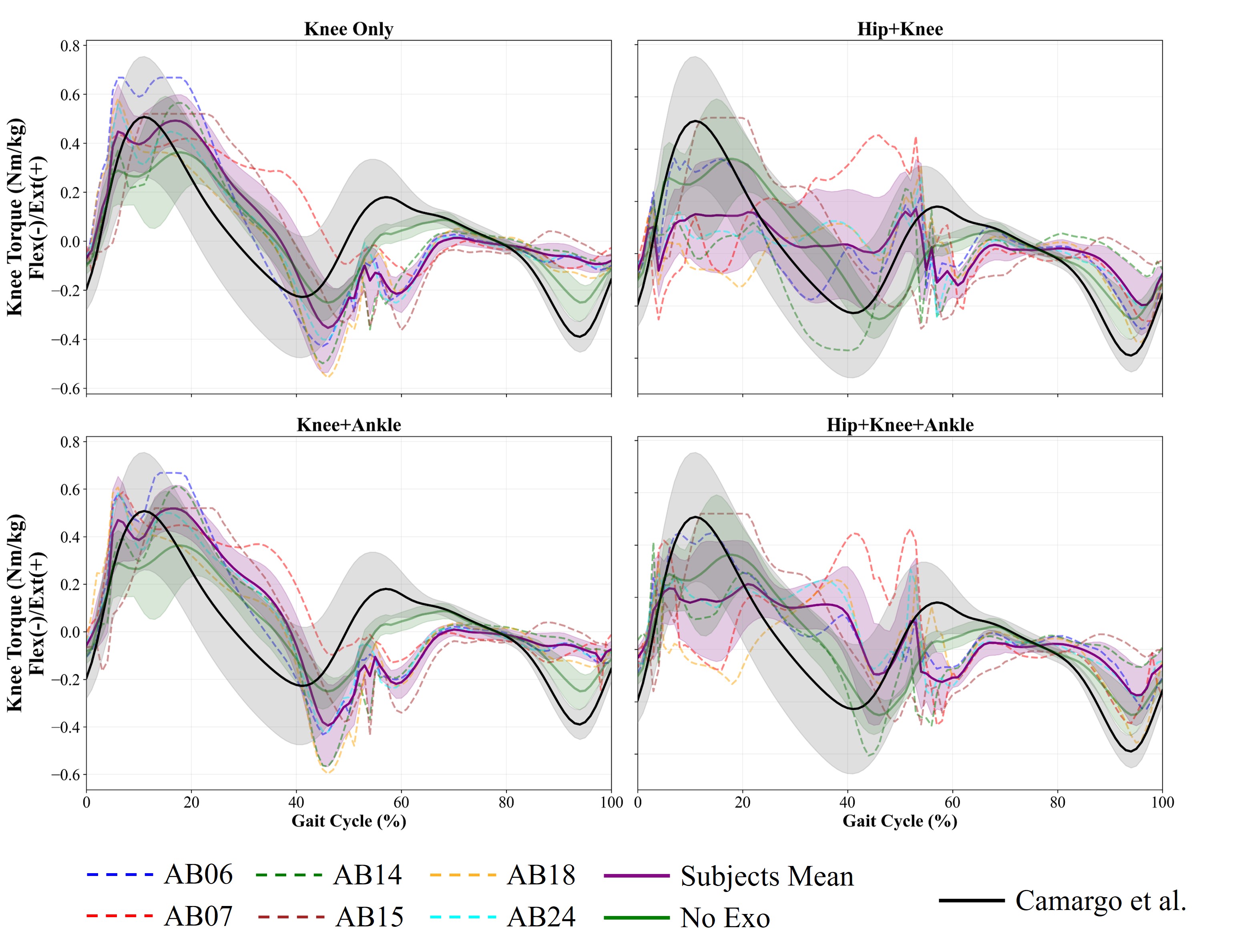}
        \caption{Predicted knee assistive torque ($50~\mathrm{Nm}$) profiles across the gait cycle for the four exoskeleton configurations (Knee Only, H+K, K+A, and H+K+A) that incorporate the knee assistance. Dashed colored lines indicate individual subject data, while the purple line (with shaded $\pm$SD) represents the subject-specific model group average. The black line shows the experimental biological knee joint torque from Camargo et al.~\cite{camargo2021comprehensive}, and the green line (with shaded $\pm$SD) depicts the mean of the unassisted condition for the subject-specific model.}
        \label{fig:knee_torques}
    \end{figure}
    
    \begin{figure}[htb!]
        \centering
        \includegraphics[width=0.98\textwidth]{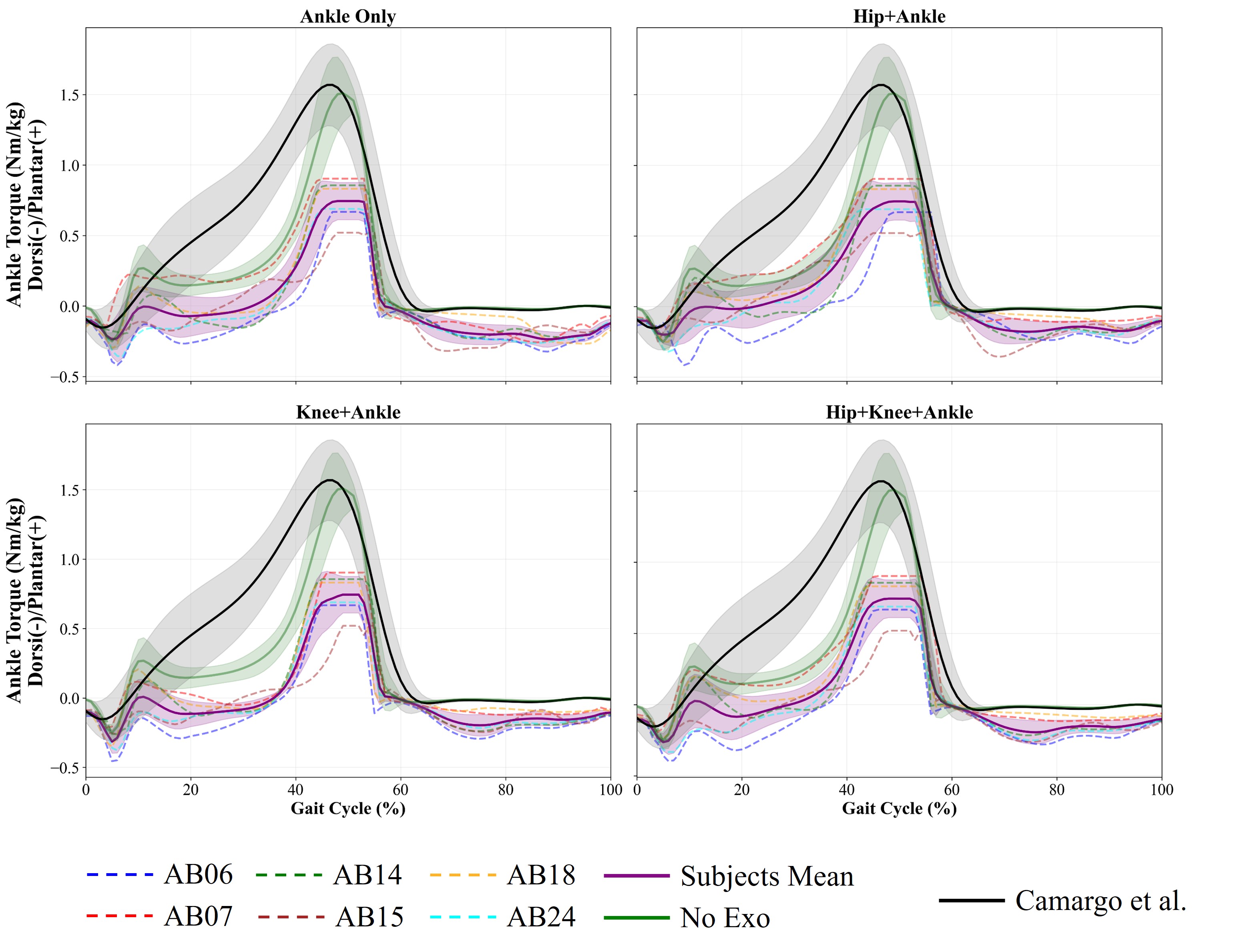}
        \caption{Predicted ankle assistive torque ($50~\mathrm{Nm}$) profiles across the gait cycle for the four exoskeleton configurations (Ankle Only, H+A, K+A, and H+K+A) that incorporate the ankle assistance. Dashed colored lines indicate individual subject data, while the purple line (with shaded $\pm$SD) represents the subject-specific model group average. The black line shows the experimental biological ankle joint torque from Camargo et al.~\cite{camargo2021comprehensive}, and the green line (with shaded $\pm$SD) depicts the mean of the unassisted condition for the subject-specific model.}
        \label{fig:ankle_torques}
    \end{figure}
    
    \begin{figure}[htb!]
        \centering
        \includegraphics[width=0.98\textwidth]{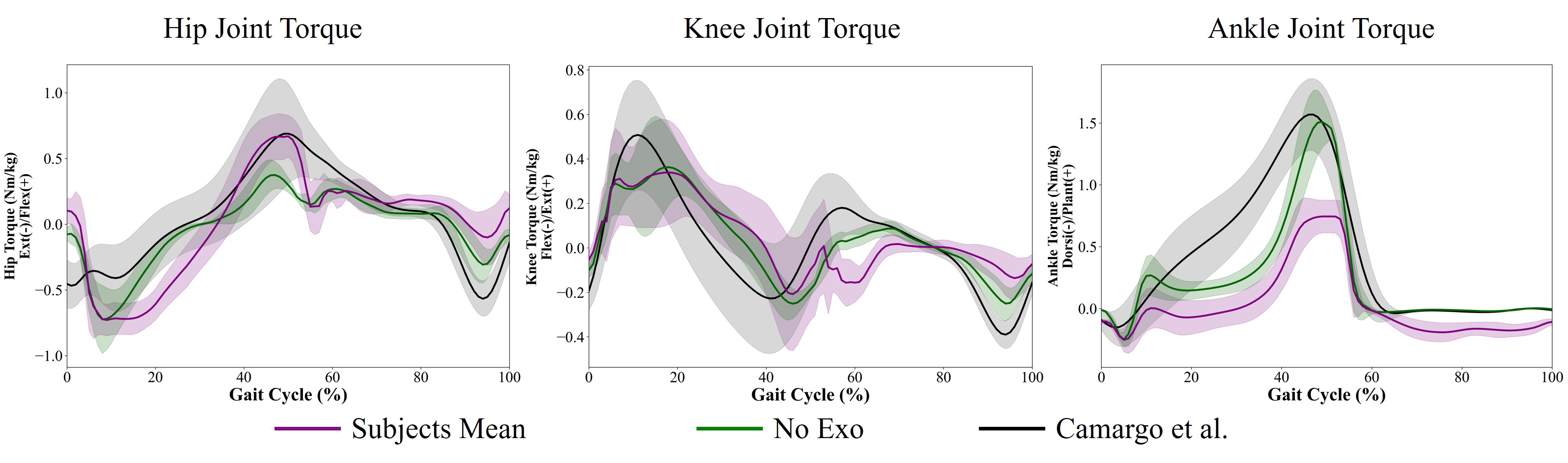}
        \caption{Overall assistive torques for the hip, knee, and ankle joints ($50~\mathrm{Nm}$) obtained across all subjects and exoskeleton assistance conditions. Purple solid lines represent the predicted mean assistive torque profiles across all subjects and conditions, with purple shaded regions denoting $\pm$1 standard deviation. The black line shows the experimental biological joint torque from Camargo et al.~\cite{camargo2021comprehensive}, and the green line (with shaded $\pm$SD) depicts the mean of the NoExo (unassisted) condition for the subject-specific model.}
        \label{fig:torques}
    \end{figure}

        \begin{figure}[htb!]
        \centering
        \includegraphics[width=0.98\textwidth]{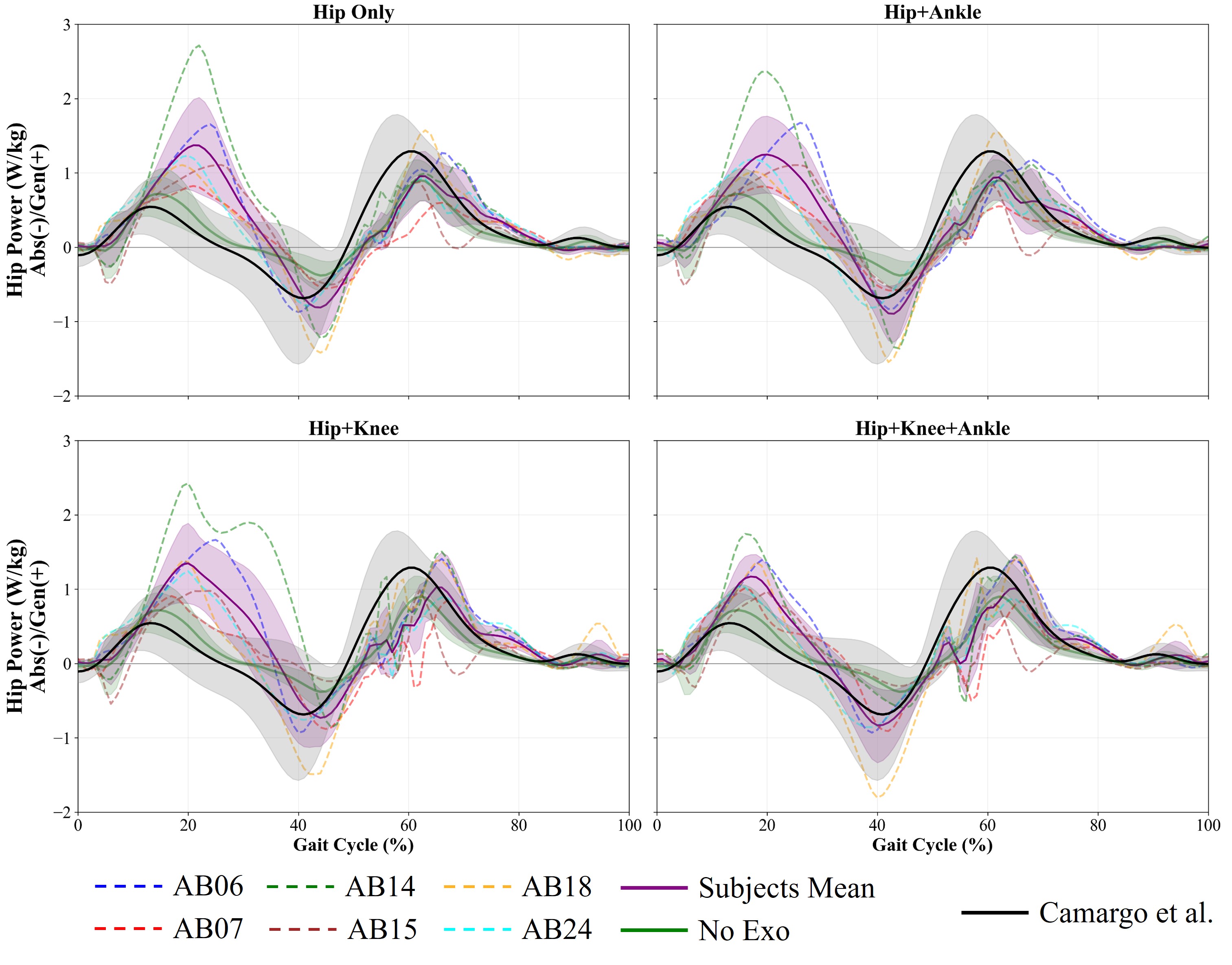}
        \caption{Predicted hip joint power under $50~\mathrm{Nm}$ assistance across the gait cycle for the four exoskeleton configurations (Hip Only, H+A, H+K, and H+K+A) that incorporate the hip. Dashed colored lines indicate individual subject data, while the purple line (with shaded $\pm$SD) represents the subject-specific model group average. The black line shows the experimental biological hip joint power profile from Camargo et al.~\cite{camargo2021comprehensive}, and the green line (with shaded $\pm$SD) depicts the mean of the NoExo condition for the subject-specific model.}
        \label{fig:hip_power}
    \end{figure}

    \begin{figure}[htb!]
        \centering
        \includegraphics[width=0.98\textwidth]{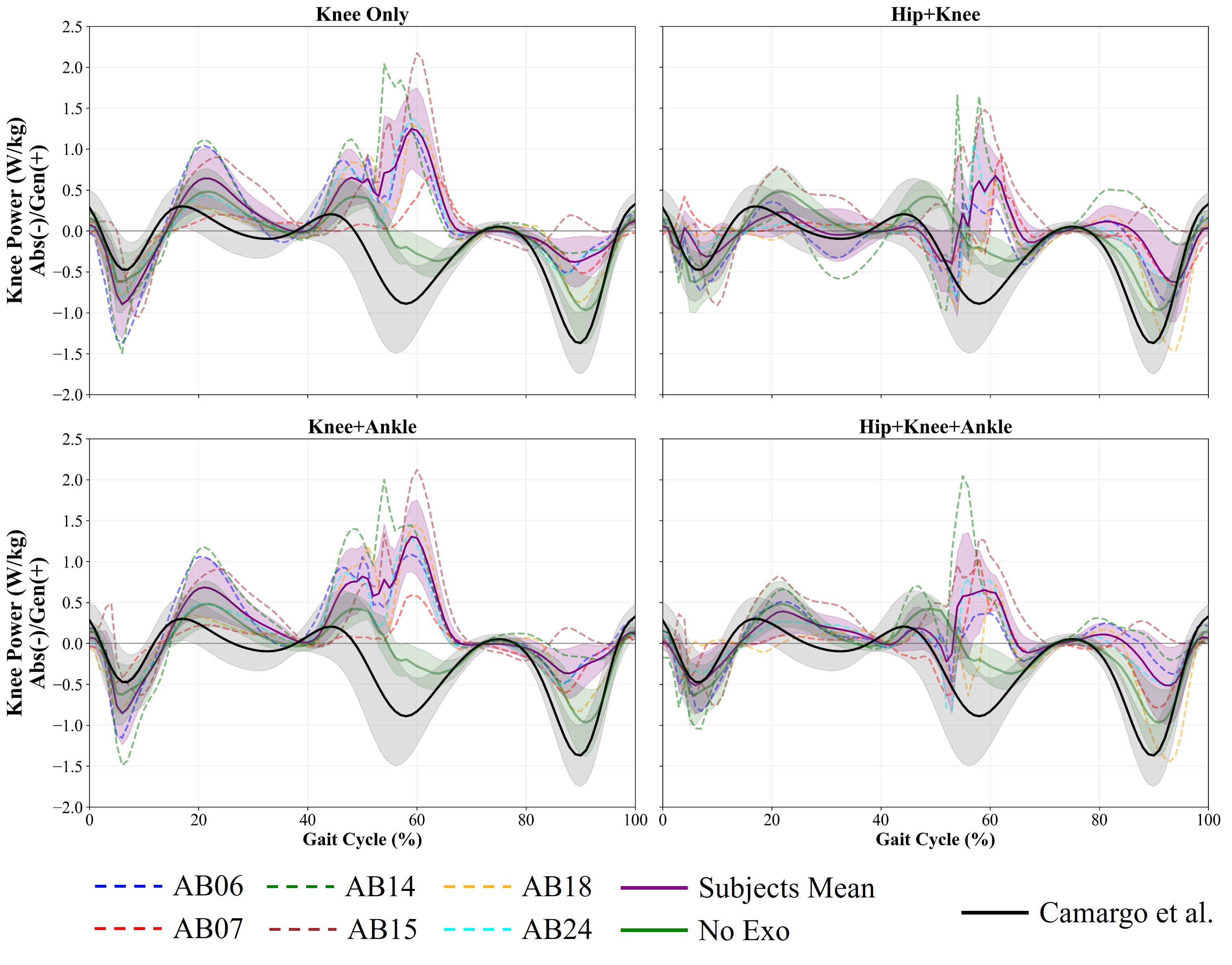}
        \caption{Predicted knee joint power under $50~\mathrm{Nm}$ assistance across the gait cycle for the four exoskeleton configurations (Knee Only, H+K, K+A, and H+K+A) that incorporate the knee. Dashed colored lines indicate individual subject data, while the purple line (with shaded $\pm$SD) represents the subject-specific model group average. The black line shows the experimental biological knee joint power profile from Camargo et al.~\cite{camargo2021comprehensive}, and the green line (with shaded $\pm$SD) depicts the mean of the NoExo condition for the subject-specific model.}
        \label{fig:knee_power}
    \end{figure}
    
    \begin{figure}[htb!]
        \centering
        \includegraphics[width=0.98\textwidth]{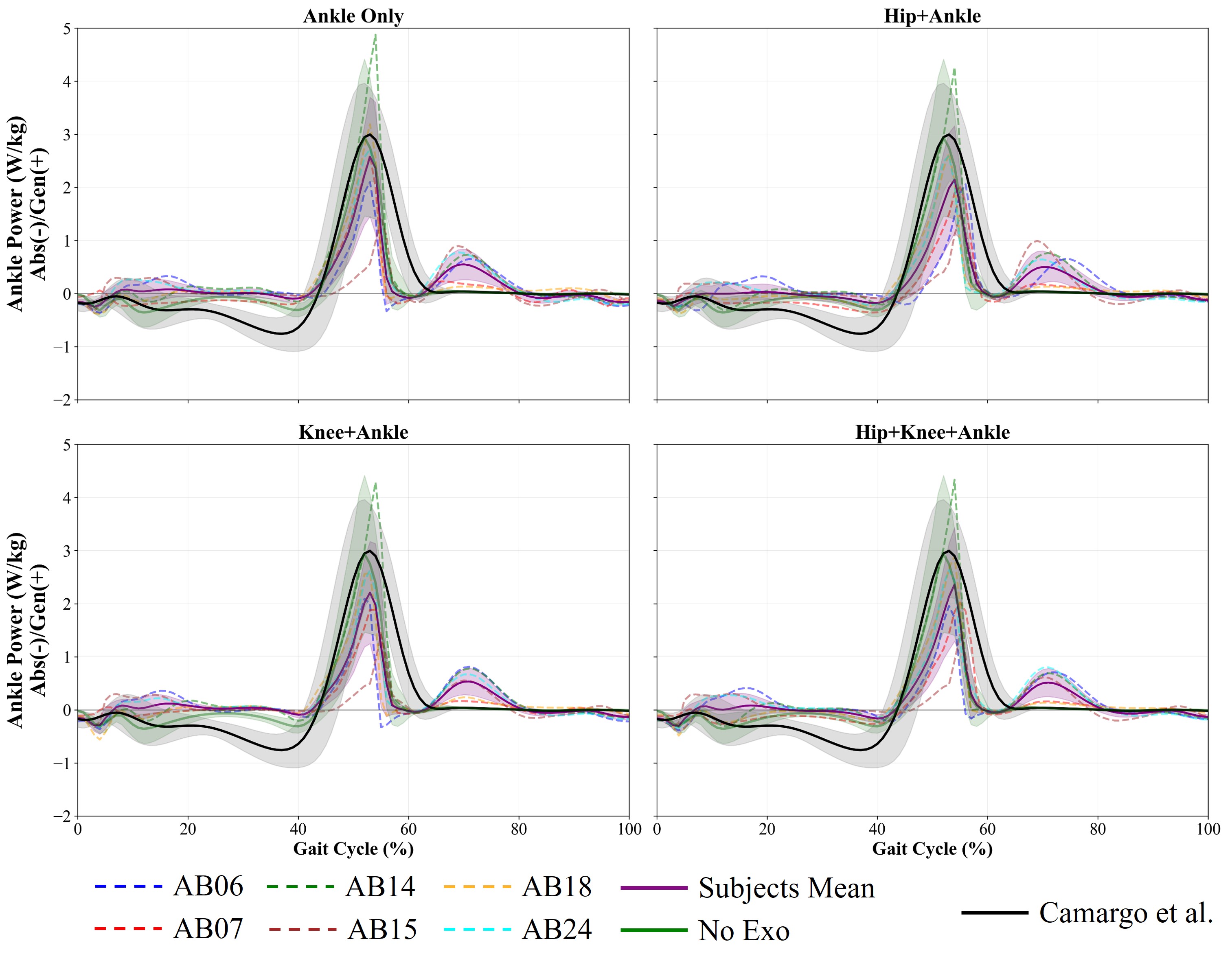}
        \caption{Predicted ankle joint power under $50~\mathrm{Nm}$ assistance across the gait cycle for the four exoskeleton configurations (Ankle Only, H+A, K+A, and H+K+A) that incorporate the ankle. Dashed colored lines indicate individual subject data, while the purple line (with shaded $\pm$SD) represents the subject-specific model group average. The black line shows the experimental biological ankle joint power profile from Camargo et al.~\cite{camargo2021comprehensive}, and the green line (with shaded $\pm$SD) depicts the mean of the NoExo condition for the subject-specific model.}
        \label{fig:ankle_power}
    \end{figure}

    \begin{figure}[htb!]
        \centering
        \includegraphics[width=0.98\textwidth]{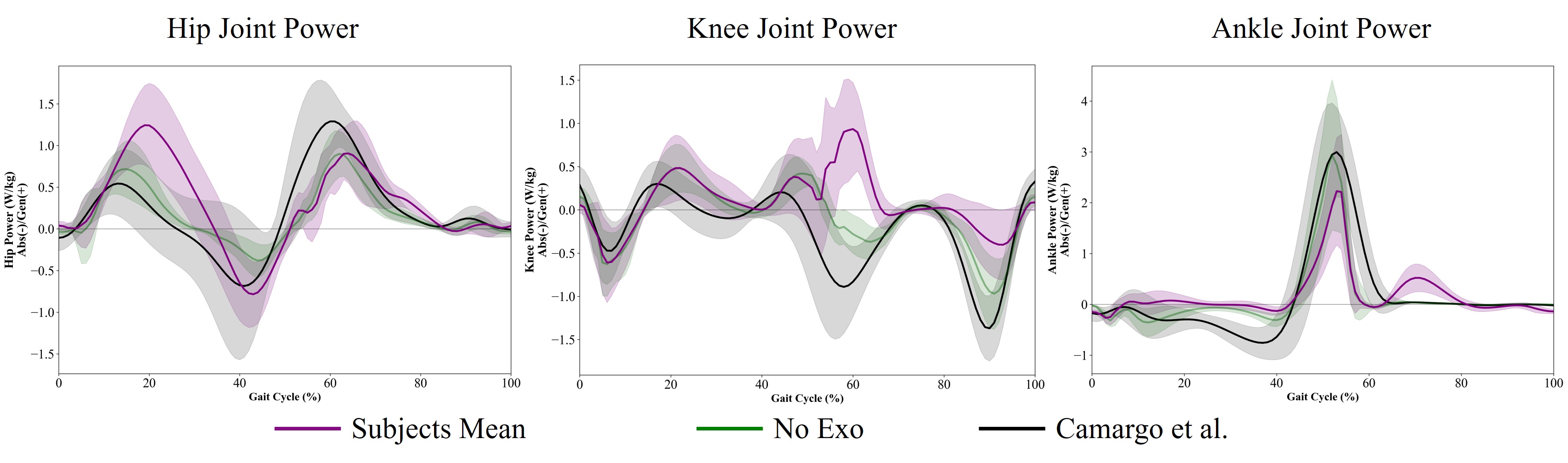}
        \caption{Overall assistive power for the hip, knee, and ankle joints ($50~\mathrm{Nm}$) obtained across all subjects and exoskeleton assistance conditions. Purple solid lines represent the predicted mean assistive power profiles across all subjects and conditions, with purple shaded regions denoting $\pm$1 standard deviation. The black line shows experimental the biological joint power from Camargo et al.~\cite{camargo2021comprehensive}, and the green line (with shaded $\pm$SD) depicts the mean of the NoExo (unassisted) condition for the subject-specific model.}
        \label{fig:power}
    \end{figure}

    \begin{figure}[htb!]
    \centering
    \includegraphics[width=0.98\textwidth]{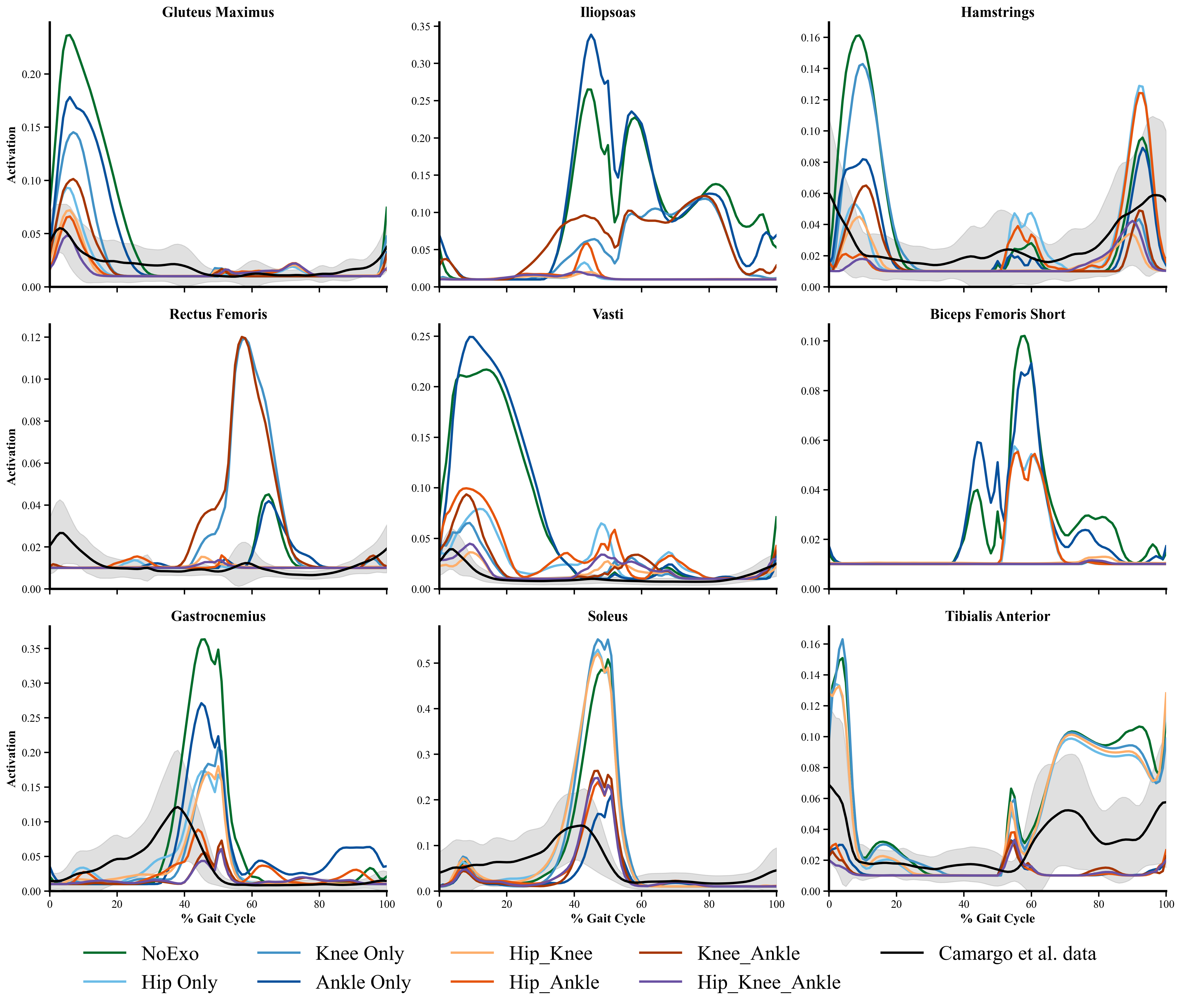}
    \caption{Muscle activation profiles across nine lower-limb muscles over the gait cycle, averaged across subjects, under NoExo and seven exoskeleton assistance conditions (Hip Only, Knee Only, Ankle Only, H+K, H+A, K+A, H+K+A) at $50~\mathrm{Nm}$ assistive torque. The blue curves in different shades show the single-joint conditions, the red shaded lines show the double-joint conditions, and the purple line shows the three-joint condition. The black line shows experimental biological muscle activations from Camargo et al.~\cite{camargo2021comprehensive} (with shaded $\pm$SD), and the green line depicts the mean of the NoExo condition for the subject-specific models. 
    *Note: The surface EMG data for IL and BFSH were not available in the dataset as they are deep muscles that are difficult to measure. In the GMax panel, the black experimental curve corresponds to GMed, as GMax data were not available.}
    \label{fig:emg_50Nm}
\end{figure}
    
\input{Power_PosNeg_Value_All}

\input{Power_PosNeg_Overall}

    \begin{figure}[htb!]
        \centering
        \includegraphics[width=1.0\textwidth]{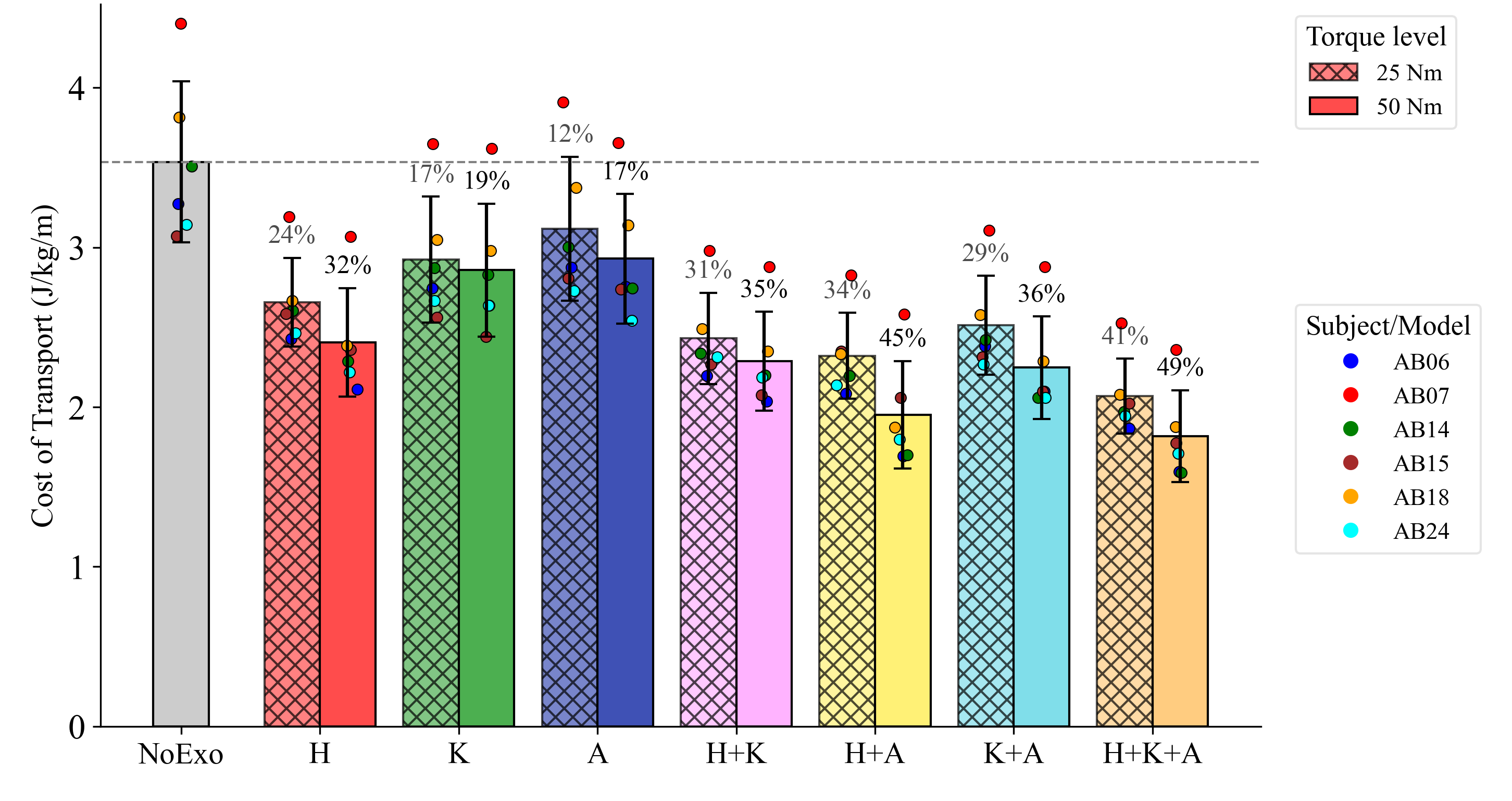}
        \caption{Mean COT values across exoskeleton assistance configurations at both $25~\mathrm{Nm}$ and $50~\mathrm{Nm}$ assistance levels. Colored dots represent individual subject-specific model results, with group means and standard deviations shown as bars.The cross-hatched bars are for $25~\mathrm{Nm}$, and the solid-color bars represent the mean COT for $50~\mathrm{Nm}$. The dashed line marks the mean unassisted NoExo COT. The percentage reduction in COT relative to the unassisted NoExo condition is shown above each bar.}
        \label{fig:COT}
    \end{figure}

    We also compiled results from predictive simulations for the assistance conditions under $25~\mathrm{Nm}$ torque, and the results are presented in Figures \ref{fig:hip_angle_25Nm}, \ref{fig:knee_angle_25Nm}, and \ref{fig:ankle_angle_25Nm}.  Comparing the joint kinematics between $50~\mathrm{Nm}$ and $25~\mathrm{Nm}$ assistive torque conditions, the mean trajectories were visually similar.
    
    The predicted joint torque profiles under the $50~\mathrm{Nm}$ exoskeleton assistance conditions are shown in Figures \ref{fig:hip_torques} through \ref{fig:ankle_torques} for individual joints and for the mean across all subjects and conditions in Figure \ref{fig:torques}. Comparing the NoExo condition with the experimental data, the predicted biological hip torque is apparently higher in (extension) magnitude during the early stance phase but lower in flexion during the terminal stance and early swing phases. During the terminal swing or landing preparation phase, the predicted hip extension torque is lower than the experimental data. The obtained torque profiles are generally consistent across individuals. The hip torque profiles (Figure \ref{fig:hip_torques}) demonstrate that the across-subject mean curve tracks reasonably close to the experimental biological torque, particularly for the Hip Only and H+K conditions. The peak hip flexion torque, occurring around 50\% of the gait cycle, shows good agreement between the across-subject mean and the experimental data in these two conditions. However, the H+K+A condition shows greater inter-subject variability, as evidenced by the wider spread of the individual curves around the across-subject mean.
    
    The knee torque profiles (Figure \ref{fig:knee_torques}) reveal some noticeable differences between the experimental data, the NoExo data, and the assisted across-subject mean. In early stance, all conditions produce an extension-torque peak, but the experimental data curves reach this peak earlier and at a slightly higher magnitude than the across-subject mean data, consistent with the earlier-loading tendency of the energy-optimized simulations noted for the hip and ankle. Through mid-to-late stance, the predicted NoExo and across-subject mean curves reach their flexion-torque trough earlier (approximately at 40-50\%) than the experimental profile, whose comparable flexion dip occurs later in the cycle.
    
    During swing, the experimental curve develops a pronounced flexion torque (approximately at 80--90\%) that the predicted profiles reproduce only weakly, remaining comparatively flat through this region, especially in the Knee Only and K+A conditions.
    
    For the ankle torques in Figure \ref{fig:ankle_torques}, the biological torque predicted in the NoExo case is similar in trend and close in peak values with the experimental reference data curves, with the exception during the middle stance phase where the predicted ankle plantarflexion torque is smaller than the experimental values. For the assistance cases, the predicted assistance torques closely mimic the trend of the predicted biological torques in the NoExo case and are clipped at $50~\mathrm{Nm}$ near the peak.  

    The predicted joint power profiles under the $50~\mathrm{Nm}$ exoskeleton assistance condition are shown in Figures \ref{fig:hip_power} through \ref{fig:power}.  
    For the hip (Figure \ref{fig:hip_power}), the assistance power is higher than that of the biological joint power, as well as the experimental data during the stance phase. 
    AB14 shows noticeably higher positive power than the other subjects; for the H+K condition at 50 Nm \acs{MPP} reaches 0.708 W/kg, roughly double the six-subject mean of 0.388 W/kg. 
    
    The knee power profiles (Figure \ref{fig:knee_power}) show notable differences between the NoExo, across-subject mean, and experimental data. In the late stance and early swing phases, the experimental and NoExo curves exhibit negative power, likely during flexion in preparation for swing. However, the across-subject mean assistance curves show positive power during this phase. Moreover, in the swing phase, the predicted assistance power is much smaller compared to the predicted (NoExo) biological joint power and the experimental data.
    
    The predicted ankle joint power (Figure \ref{fig:ankle_power}) has very consistent curves across conditions, where it roughly tracks the predicted biological power as well as the experimental data curves. While small negative power values (-0.040 W/kg for the H+A condition) are observed during the early and middle stance phases in the predicted profiles (Figure \ref{fig:ankle_power}), their magnitude is minimal when contrasted with the more obvious negative power phases in the experimental data. 

    Table \ref{tab:Power_PosNeg_Value_All} lists the power values of \acs{MPP}, \acs{MNP}, and \acs{PPR} for hip, knee, and ankle joints for both $25~\mathrm{Nm}$ and $50~\mathrm{Nm}$ assistance conditions, and the metabolic reduction values compared to the Baseline value of 3.07 W/Kg. Table \ref{tab:Power_PosNeg_Overall} presents the mean power magnitudes for both assistance levels, each averaged from all corresponding joint assistance conditions. 
    Comparing single-joint assistance at $50~\mathrm{Nm}$, mean positive power differed substantially across joints (Table~\ref{tab:Power_PosNeg_Value_All}): the hip delivered the highest \acs{MPP} (0.375 W/kg), followed by the knee (0.243 W/kg) and ankle (0.220 W/kg).
    
    Across both tables, the hip and ankle consistently have higher values of positive power, while the knee has considerably higher negative power. At $50~\mathrm{Nm}$, the mean \acs{PPR} is highest for the ankle (0.861) and hip (0.820) and lowest for the knee (0.672), indicating that roughly 86\% and 82\% of the ankle and hip assistive power is delivered as positive work, versus only about 67\% at the knee (Table~\ref{tab:Power_PosNeg_Overall}). This lower knee ratio is driven by its relatively large negative power: at $50~\mathrm{Nm}$, the knee \acs{MNP} reaches -0.082 W/kg on average, comparable to the hip's -0.079 W/kg but paired with a much smaller positive component, so a greater proportion of knee work is negative. The effect is most pronounced in the H+K condition, where the knee \acs{PPR} drops to 0.442 with an \acs{MPP} of just 0.073 W/kg (Table~\ref{tab:Power_PosNeg_Value_All}), reflecting the flipped-sign flexion assistance the optimizer selects near pre-swing.

    Increasing assistance from $25~\mathrm{Nm}$ to $50~\mathrm{Nm}$ raised mean positive power at every joint, but unevenly. Hip \acs{MPP} increased from 0.267 to 0.359 W/kg and ankle \acs{MPP} from 0.124 to 0.205 W/kg, whereas knee \acs{MPP} changed only from 0.163 to 0.168 W/kg (Table~\ref{tab:Power_PosNeg_Overall}). A similar pattern appeared in the peak and RMS assistive torques: $\tau_{MAX}$ increased from 0.372 to 0.729 Nm/kg at the hip and from 0.372 to 0.745 Nm/kg at the ankle, but only from 0.310 to 0.339 Nm/kg at the knee.

    Figure \ref{fig:COT} illustrates the mean COT and percentage COT reductions across all assistance configurations at both $25~\mathrm{Nm}$ and $50~\mathrm{Nm}$ torque levels, with per-subject values reported in Table \ref{tab:COT_subjectwise}. Under all assistance conditions, the COT was reduced compared with the baseline unassisted (NoExo) case. Across both torque levels, the greatest mean reductions in COT were observed in the multi-joint hip-knee-ankle ($\text{H+K+A}$) condition with reductions of $41.20\%$ and $48.50\%$ respectively. For the H+K+A condition at $50~\mathrm{Nm}$, the predicted COT reductions ranged between $42.22\%$ for AB15 and $54.65\%$ for AB14, with all six subject-specific models showing reductions that are greater than $42\%$. The smallest mean reductions occurred in the $25~\mathrm{Nm}$ ankle-only ($\text{A}$) condition at $11.85\%$. Among single-joint conditions, the hip was consistently the most effective and the ankle the least, mirroring the joint-power ordering noted above. At the torque level of $25~\mathrm{Nm}$, individual subject reductions ranged from $8.64\%$ ($\text{AB15}$) to $45.54\%$ ($\text{AB18}$). Increasing the torque magnitude to $50~\mathrm{Nm}$ resulted in larger reductions, ranging from $10.84\%$ ($\text{AB15}$) to $54.65\%$ ($\text{AB14}$).  

    Beyond the general increase with torque and joint count, most of the predicted benefit came from hip and ankle assistance. The hip--ankle ($\text{H+A}$) configuration reached $44.60\%$ at $50~\mathrm{Nm}$, the strongest two-joint result and close to the full hip--knee--ankle ($\text{H+K+A}$) condition.

    When increasing the assistance torque from $25~\mathrm{Nm}$ to $50~\mathrm{Nm}$, the largest change in mean savings occurred in the hip-ankle ($\text{H+A}$) condition, where mean COT reduction rose from $33.95\%$ to $44.60\%$ (an absolute increase of $10.65\%$). In contrast, the knee-only (K) condition demonstrated less sensitivity to torque magnitude; increasing the torque from $25~\mathrm{Nm}$ to $50~\mathrm{Nm}$ only improved mean metabolic savings from $17.20\%$ to $19.17\%$, representing a minor absolute change of approximately $2\%$.
    At $50~\mathrm{Nm}$, COT reduction differed significantly among both
the single-joint and two-joint configurations (Holm-adjusted
$p=0.0166$ for both comparisons). However, adding knee assistance to
H+A did not provide a statistically significant additional reduction
(Holm-adjusted $p=0.0625$; Appendix
Table~\ref{tab:cot_minimal_statistics}).
    
    Figure \ref{fig:emg_50Nm} presents the mean muscle activation profiles for the nine modeled lower-limb muscles under the NoExo baseline and all seven assistance conditions at $50~\mathrm{Nm}$, with the corresponding $25~\mathrm{Nm}$ profiles shown in Figure \ref{fig:emg_25Nm}. Assistance generally reduced activation in muscles spanning the assisted joint. Hip assistance reduced mean GMAX activation by approximately $59.75\%$ for Hip Only and $68.80\%$ for H+K+A, while IL activation decreased by approximately 87\% across hip-assisted conditions. Conditions without hip assistance left these muscles near the NoExo profile.
    
    Knee assistance primarily affected the knee extensors and flexors. The VAS are strongly active during stance in the NoExo condition and showed a large reduction (about $68.75\%$ for Knee Only) whenever the knee was assisted. Conversely, the RF displayed the opposite trend: its mid-cycle activation increased in the knee-assisted conditions (Knee Only and K+A), rising to a distinct peak near 50\% of the gait cycle that was absent at baseline.
    
    Ankle assistance reduced plantarflexor activation, although the two plantarflexors responded differently. The SOL showed a substantial reduction with ankle assistance (about $56.42\%$ for Ankle Only), whereas the GAS was reduced by only $10.91\%$ with Ankle Only, by $49.14\%$ with Knee Only, and by $72.90\%$ with K+A. The TA was essentially unaffected during its early-stance burst but showed reduced late-swing activation. Notably, several conditions produced plantarflexor activations exceeding the experimental Camargo et al.~\cite{camargo2021comprehensive} envelope around the stance-to-swing transition, particularly for the SOL in hip- and knee-assisted conditions.
    
    Comparing the two torque levels, the $25~\mathrm{Nm}$ profiles (Figure \ref{fig:emg_25Nm}) followed the same qualitative pattern as the $50~\mathrm{Nm}$ profiles but with consistently smaller reductions. For example, GMAX activation under hip assistance fell by about $37.21\%$ at $25~\mathrm{Nm}$ versus $59.75\%$ at $50~\mathrm{Nm}$, reflecting the greater biological offloading available at higher assistive torque.

    \section{Discussion}\label{sec:discussion}

    This study compared all combinations of idealized hip, knee, and ankle assistance at two torque limits across six anthropometrically scaled musculoskeletal models. Three findings were consistent within the optimization framework. First, hip assistance produced the greatest single-joint reduction in predicted COT. Second, H+A captured most of the predicted benefit of H+K+A, indicating very small gains from adding knee assistance. Third, knee assistance exhibited a distinct positive-power strategy and comparatively little sensitivity to the higher torque limit. The following discussion interprets these outcomes and their implications; the supporting values are reported in Section \ref{sec:results} and Tables \ref{tab:COT_subjectwise} and \ref{tab:Power_PosNeg_Overall}.

    The predicted trajectories reproduced the principal sagittal-plane features of the experimental gait data, particularly at the knee, but systematic discrepancies remained at the hip and ankle (Figure \ref{fig:hip_angle}-\ref{fig:ankle_angle}). These discrepancies are informative rather than incidental. The smaller simulated hip extension reflects the compromise between the tracking and effort terms; the underpredicted ankle plantarflexion is consistent with the model's simplified foot–ground contact and absent toe joint; and the elevated early-stance hip torque plausibly follows from the same missing passive damping and soft-tissue absorption. Because these effects act on the unassisted baseline, they constrain claims of subject-level biofidelity and may propagate into the optimized assistance magnitudes.

    Kinematic profiles were nearly invariant to the torque bounds, whereas muscle activation and joint power changed more clearly (Table \ref{tab:Power_PosNeg_Overall}). Within this formulation, the added actuator capacity therefore redistributed mechanical demand rather than producing a different sagittal-plane gait pattern. This interpretation is limited by the tracking term, which constrains kinematic adaptation toward the reference gait.
         
    The predicted-versus-experimental torque differences localize where the framework is least reliable. The overpredicted early-stance hip-extension torque, set against the closer knee agreement (Figure \ref{fig:torques}), follows from an effort-minimizing objective that seeks an idealized pattern lacking the passive damping and soft-tissue absorption of human locomotion, compounded by simplified contact dynamics that amplify loading-response demand. The ankle deficit through mid-stance, largest near the plantarflexion peak, again points to the simplified foot–ground contact and missing toe joint. Because an inflated baseline can inflate the assistance the model deems necessary, the temporal patterns are best treated as controller hypotheses while their absolute magnitudes await experimental validation and more biofidelic contact and energy-absorption models.
    
    The reversal in the knee power pattern is a notable feature of the optimized assistance. During normal gait, knee joint power is predominantly negative as the knee flexors decelerate the extending knee before heel-strike~\cite{li2009development}. By contrast, the optimized assistance delivered positive knee power in the pre-swing window, and the assisted knee PPR rose well above both the predicted and experimental unassisted values (Table \ref{tab:Power_PosNeg_Overall}). Rather than scaling the biological extension torque, energy-optimal knee assistance redirects the joint toward positive pre-swing power. Franks et al. ~\cite{franks2021comparing} observed the same qualitative strategy, which supports reading this as a genuine energetic solution rather than an artifact of the formulation.

   
    Raising the peak assistive torque increased the mean \acs{COT} reduction in every configuration, but the additional benefit depended strongly on the assisted joints (Figure \ref{fig:COT}). H+A was the most responsive multi-joint configuration, whereas Knee Only and H+K changed relatively little. The joint-power and torque metrics support this pattern: the higher bound produced substantial increases in hip and ankle \acs{MPP} and $\tau_{MAX}$, while the corresponding knee measures changed only modestly (Table~\ref{tab:Power_PosNeg_Overall}). Thus, the benefit of greater torque capacity was joint-dependent and showed diminishing returns, particularly at the knee.
  
    Our prediction revealed a hip-then-knee-then-ankle ordering for single-joint assistance (Figure \ref{fig:COT}), and the joint-power analysis explains why: the hip delivered the highest mean positive power (Table \ref{tab:Power_PosNeg_Overall}) and therefore the greatest positive mechanical work available for an external device to replace. The ankle yielded the smallest reduction. This is consistent with results from Seo et al.~\cite{seo2016fully}, who reported roughly twice the metabolic benefit from hip versus ankle exoskeletons and attributed it to the lower efficiency of the hip muscle–tendon units. At the ankle, the compliant gastrocnemius–soleus–Achilles complex stores and returns elastic energy during push-off, leaving little costly muscle work to offload, whereas the hip relies on active concentric work that is more readily replaced. The one departure from experimental expectation, knee assistance slightly exceeding ankle assistance, inverting the ankle-over-knee ordering usually observed~\cite{franks2021comparing} is most plausibly attributed to the same simplified ankle–foot mechanics and the model's positive-power knee strategy, and should be read with that caveat.
    
    Combining joints produced additive but sub-linear gains. H+A recovered most of the full-configuration benefit, so adding knee actuation yielded only a modest further reduction (Figure \ref{fig:COT}). This agrees in direction with the experimental results of Franks et al.~\cite{franks2021comparing}, who likewise found that the benefits of individually optimized joints did not sum linearly when combined, reflecting mechanical and control interactions among joints. The absolute reductions predicted here were larger than those measured experimentally, as expected from an idealized, energy-optimized model that discourages co-contraction, full muscle redundancy, device limitations, and constraints on human adaptation. Nevertheless, the relative results identify H+A as a promising compromise between predicted metabolic benefit and actuator complexity.

    The muscle activation results reinforce the joint-power and COT findings and clarify how assistance reduces metabolic cost: applied torque offloads the biological muscles spanning the assisted joint, shifting mechanical demand onto the exoskeleton. This load-sharing was joint-specific and localized, with each condition suppressing the agonists of the joint it targeted while leaving muscles at unassisted joints near baseline. The strong suppression of the GMAX and IL under hip assistance is consistent with the hip producing the largest single-joint COT reduction, as these are among the most metabolically costly muscles during walking.

    Some responses reflect redistribution of effort rather than uniform reduction. The RF increased its mid-cycle activation in knee-assisted conditions even as the VAS decreased, being recruited to meet demands that arise when the exoskeleton alters the knee moment; this peak, though conspicuous against a near-zero baseline, remains modest in absolute terms ($\sim$ 0.12) relative to the primary movers such as the SOL ($\sim$ 0.51) and GMAX ($\sim$0.24). Similarly, the biarticular GAS was only weakly reduced by ankle assistance alone but fell sharply once the knee was co-assisted, a coupling that helps explain why multi-joint conditions achieve muscle-level savings beyond single-joint assistance.

    Two activation results warrant restraint. The BFSH showed low activation and little systematic change, but was difficult to validate: as a deep muscle beneath the long head it cannot be recorded selectively with surface electrodes, and even fine-wire placement risks the overlying long head~\cite{hass2026ultrasound}, with thigh surface–fine-wire comparisons finding no adequate surface proxy~\cite{kingston2018representing}; agreement or discrepancy for this muscle should not be over-interpreted. Separately, some plantarflexor activations, notably soleus, met or exceeded the experimental Camargo et al.~\cite{camargo2021comprehensive} envelope in certain conditions. With only nine muscles per limb, no co-contraction penalty beyond the effort term, and rigid tendons that omit elastic energy storage, the model may over-recruit individual plantarflexors to meet push-off demands a compliant biological ankle would satisfy more economically, consistent with the underprediction of absorption power noted earlier. These factors temper interpretation of absolute magnitudes, though the relative, joint-specific offloading patterns are physiologically plausible.
    


    The metabolic reduction values reported in this study should be interpreted with caution, as they are based on a musculoskeletal model with a limited number of muscles and an energy-optimized gait pattern. The model does not explicitly capture factors such as muscle co-contraction and the full redundancy of human motor control. Consequently, the simulated baseline gait may underestimate the metabolic cost of real human locomotion and may not fully represent the capacity for metabolic savings through exoskeleton assistance. Thus, the absolute magnitude of the predicted metabolic reductions may differ from that observed experimentally, even though the relative differences among assistance configurations may provide useful insight into joint-specific sensitivity to assistance. Future work should examine these effects using more comprehensive neuromusculoskeletal models and, critically, experimental measurements from human participants.

    An additional consideration is the extent to which the predicted assistance strategy must be personalized to individual users. Recent human-in-the-loop experiments suggest that individuals may not always necessarily require fully customized torque profiles to achieve metabolic benefits. For example, Bryan et al.~\cite{bryan2024pilot} showed that different subjects could benefit similarly from a generalized torque pattern, with torque timing being more consistent between participants than torque magnitude. This finding supports the potential for broadly applicable assistance strategies, but it does not eliminate the need to assess the robustness of simulation-derived predictions. In particular, predictive simulation outcomes can be sensitive to the formulation of the optimization objective; Nikoo et al.~\cite{nikoo2022careful} demonstrated that seemingly minor changes in cost terms or their relative weighting can substantially affect predicted locomotor behavior. Therefore, the torque profiles and metabolic benefits identified here should be viewed as model-dependent predictions rather than direct estimates of human responses. Experimental validation, such as the biomechanical measurements reported by Firouzi et al.~\cite{firouzi2025biomechanical}, will be essential for determining how reliably these simulation-based predictions translate to human locomotion and for identifying which aspects of the predicted assistance strategy are robust across individuals.

    \section{Conclusions}\label{sec:conclusions}
   This study used subject-specific predictive simulations to identify energy-optimal lower-limb exoskeleton assistance across single- and multi-joint configurations at $25~\mathrm{Nm}$ and $50~\mathrm{Nm}$. The simulated kinematics and muscle activations were physiologically plausible and reproduced the expected joint-specific muscle offloading with assistance. 
   Across six anthropometrically scaled models, the predictive simulations consistently favored hip assistance among single-joint configurations and H+A among two-joint configurations. At $50~\mathrm{Nm}$, H+A recovered most of the predicted COT benefit of full H+K+A assistance, indicating a limited incremental gain from adding knee actuation in this formulation (Figure \ref{fig:COT}). Knee assistance also differed mechanically from hip and ankle assistance: it provided positive power near pre-swing in opposition to biological power and showed little sensitivity to the higher torque bound.
   Overall, these findings show that predictive simulations can offer useful joint-level guidance for exoskeleton design. Future work should incorporate more biofidelic models and experimental validation to further assess and translate these strategies to hardware.

    \section*{CRediT authorship contribution statement}
    \textbf{N. Ratnakumar}: Writing – original draft, Simulations, Visualization, Data curation; \textbf{M. Tohfafarosh}: Writing – review \& editing, Visualization, Data curation.
    \textbf{X. Zhou}: Writing – review \& editing, Funding acquisition, Supervision, Conceptualization.
    
    \section*{Declaration of competing interest}
    The authors declare that they have no known competing financial interests or personal relationships that could have appeared to influence the work reported in this paper.
    
    \section*{Ethics statement}\label{sec:Ethics}
This study did not require ethical approval as it exclusively utilized publicly available data and did not involve human participants, animal experiments, or access to confidential patient information.

    \section*{Funding}\label{sec:acknowledgements}
This work was supported by NSF Award \#2524089 and NIDILRR funded RERC Grant 90REGE0025-01-00.

    \section*{Declaration of generative AI and AI-assisted technologies in the manuscript preparation process}\label{sec:AI_use}
During the preparation of this work, the author(s) used Claude for language and grammar editing. The author(s) reviewed and edited the output as needed and take full responsibility for the content of the published article. 

    \clearpage

    \bibliography{mybibfile}

    \clearpage

    \input{appendix}

\end{document}

%% file: myglossary.tex
\usepackage[nonumberlist,nogroupskip]{glossaries}

\newglossaryentry{7730}
{
name={ISO~7730:2005},
description={ISO~7730:2005 is a thermal comfort standard developed by ISO}
}

\newglossaryentry{55}
{
name={ASHRAE~55--2020},
description={ASHRAE~55--2020 is a thermal comfort standard developed by ANSI and ASHRAE}
}

\makenoidxglossaries

%% file: myacronyms.tex

\section*{Nomenclature}
\renewcommand{\baselinestretch}{0.75}\normalsize
\renewcommand{\aclabelfont}[1]{\textsc{\acsfont{#1}}}
\begin{acronym}[longest] 

\acro{COT}[COT]{Cost of Transport \acroextra{, J/kg/m}}
\acro{MPP}[MPP]{Mean Positive Power\acroextra{, W/kg}}
\acro{MNP}[MNP]{Mean Negative Power\acroextra{, W/kg}}
\acro{DOF}[DOF]{Degree of Freedom}
\acro{SCONE}[SCONE]{Simulated Controller OptimizatioN Environment}
\acro{BMI}[BMI]{Body Mass Index}
\acro{HIL}[HIL]{Human-in-the-loop}
\acro{CMC}[CMC]{Computed Muscle Control}
\acro{F-max}[$F_{max}$]{Maximum Isometric Muscle Forces\acroextra{, N}}
\acro{EMG}[EMG]{Electromyography}
\acro{F-max-G}[$F_{max}^{G}$]{Generic Model Maximum Isometric Muscle Forces\acroextra{, N}}
\acro{H}[$H$]{Model Height\acroextra{, m}}
\acro{H-G}[$H^G$]{Generic Model Height\acroextra{, m}}
\acro{J-T}[$J_T$]{Cost Function (Total Cost)}
\acro{J-MER}[$J_{MER}$]{Metabolic Cost Term}
\acro{J-COM}[$J_{COM}$]{Center of Mass Acceleration Cost Term}
\acro{J-Joint}[$J_{Joint}$]{Joint Acceleration Cost Term}
\acro{J-Effort}[$J_{Effort}$]{Muscle Effort Cost Term}
\acro{J-Track}[$J_{Track}$]{Tracking Term}
\acro{M}[$M$]{Model Mass\acroextra{, kg}}
\acro{M-G}[$M^G$]{Generic Model Mass\acroextra{, kg}}
\acro{MSK}[MSK]{Musculoskeletal}
\acro{R-MSK}[$R_{MSK}$]{Proportion of Model Mass Composed of Muscle}
\acro{R-MSK-G}[$R_{MSK}^{G}$]{Proportion of Model Mass Composed of Muscle in the Generic Model}
\acro{PPR}[PPR]{Positive Power Ratio}
\acro{COM}[COM]{Center of Mass}

\end{acronym}
\renewcommand{\baselinestretch}{1}\normalsize

%% file: subject_demo.tex
\begin{table}[h!]
        \centering
        \caption{Subject Demographics and Tracking Data Walking Speeds}
        \label{tab:subject_demographics}
        \begin{tabular}{lcccccc}
            \hline
            Subject ID & Age & Height (m) & Weight (kg) & BMI & Sex & Tracking Speed (m/s) \\
            \hline
            AB06    & 20 & 1.80 & 74.84 & 23.10 & M & 1.42 \\
            AB07    & 20 & 1.65 & 55.34 & 20.33 & M & 0.92 \\
            AB14    & 22 & 1.52 & 58.41 & 25.28 & F & 1.68 \\
            AB15    & 21 & 1.78 & 96.16 & 30.35 & M & 1.21 \\
            AB18    & 19 & 1.80 & 60.13 & 18.56 & F & 1.46 \\
            AB24    & 21 & 1.73 & 72.57 & 24.25 & F & 1.37 \\
            \hline
            Average & 20.5 & 1.71 & 69.58 & 23.64 & -- & 1.34 \\
            SD     & 1.05 & 0.11 & 15.23 & 4.13  & -- & 0.26 \\
            Generic & --   & 1.70 & 62.00 & 21.45 & -- & 1.33 \\
            \hline
        \end{tabular}
    \end{table}

%% file: Power_PosNeg_Value_All.tex
\begin{table}[htb!]
    \centering
    \small
    \setlength{\tabcolsep}{4pt}
    \renewcommand{\arraystretch}{1.2}
    \caption{Subjects' mean power of individual joints under $25~\mathrm{Nm}$ and $50~\mathrm{Nm}$ assistive torques and different combinations of joint assistance. The corresponding metabolic reduction is also presented.}
    \label{tab:Power_PosNeg_Value_All}
    \begin{tabular}{|c|c|c|c|c|c|c|}
        \hline
        \textbf{Joint} & \shortstack{\textbf{Torque} \\ \textbf{(Nm)}} & \textbf{Condition} & \shortstack{\textbf{MPP} \\ \textbf{(W/kg)}} & \shortstack{\textbf{MNP} \\ \textbf{(W/kg)}} & \textbf{PPR} & \shortstack{\textbf{Metabolic} \\ \textbf{Reduction (W/kg)}} \\ \hline
        
        \multirow{8}{*}{Hip} & \multirow{4}{*}{25} & Hip Only & 0.279 & -0.052 & 0.843 & 1.032 \\ \cline{3-7}
         &  & H+A & 0.274 & -0.053 & 0.838 & 1.442 \\ \cline{3-7}
         &  & H+K & 0.276 & -0.049 & 0.849 & 1.345 \\ \cline{3-7}
         &  & H+K+A & 0.248 & -0.062 & 0.800 & 1.778 \\ \cline{3-7}
 \cline{2-7}
        
         & \multirow{4}{*}{50} & Hip Only & 0.375 & -0.081 & 0.822 & 1.323 \\ \cline{3-7}
         &  & H+A & 0.371 & -0.085 & 0.814 & 1.869 \\ \cline{3-7}
         &  & H+K & 0.388 & -0.068 & 0.851 & 1.464 \\ \cline{3-7}
         &  & H+K+A & 0.316 & -0.095 & 0.769 & 2.050 \\ \cline{3-7}
 \hline
        
        \multirow{8}{*}{Knee} & \multirow{4}{*}{25} & Knee Only & 0.209 & -0.089 & 0.701 & 0.790 \\ \cline{3-7}
         &  & H+K & 0.108 & -0.090 & 0.545 & 1.345 \\ \cline{3-7}
         &  & K+A & 0.210 & -0.089 & 0.702 & 1.285 \\ \cline{3-7}
         &  & H+K+A & 0.136 & -0.080 & 0.630 & 1.778 \\ \cline{3-7}
 \cline{2-7}
        
         & \multirow{4}{*}{50} & Knee Only & 0.243 & -0.103 & 0.702 & 0.868 \\ \cline{3-7}
         &  & H+K & 0.073 & -0.092 & 0.442 & 1.464 \\ \cline{3-7}
         &  & K+A & 0.270 & -0.091 & 0.748 & 1.596 \\ \cline{3-7}
         &  & H+K+A & 0.128 & -0.084 & 0.604 & 2.050 \\ \cline{3-7} 
 \hline
        
        \multirow{8}{*}{Ankle} & \multirow{4}{*}{25} & Ankle Only & 0.134 & -0.038 & 0.779 & 0.493 \\ \cline{3-7}
         &  & H+A & 0.118 & -0.040 & 0.747 & 1.442 \\ \cline{3-7}
         &  & K+A & 0.126 & -0.029 & 0.813 & 1.285 \\ \cline{3-7}
         &  & H+K+A & 0.130 & -0.033 & 0.798 & 1.778 \\ \cline{3-7}
 \cline{2-7}
        
         & \multirow{4}{*}{50} & Ankle Only & 0.220 & -0.035 & 0.863 & 0.720 \\ \cline{3-7}
         &  & H+A & 0.195 & -0.040 & 0.830 & 1.869 \\ \cline{3-7}
         &  & K+A & 0.209 & -0.028 & 0.846 & 1.596 \\ \cline{3-7}
         &  & H+K+A & 0.202 & -0.038 & 0.842 & 2.050 \\ \cline{3-7}
 \hline
    \end{tabular}
\end{table}

%% file: Power_PosNeg_Overall.tex
\begin{table}[h]
    \centering
    \small
    \renewcommand{\arraystretch}{1.2}
    \caption{Mean Joint Power Magnitudes under 25 Nm and 50 Nm assistive torques (averaged from all corresponding joint assistance cases). All PPR values were computed from the instantaneous joint power time series using Eq. \ref{PPR_Equation}. For comparison, the unassisted biological knee PPR, computed the same way, was 0.352 for the predicted NoExo condition and 0.162 for the experimental data.}
    \label{tab:Power_PosNeg_Overall}
    \begin{tabular}{|c|c|r|r|r|r|r|}
   
     \hline
        \textbf{Joint} & \shortstack{\textbf{$\tau$} \\ \textbf{(Nm)}} &  \shortstack{\textbf{MPP} \\ \textbf{(W/kg)}} & \shortstack{\textbf{MNP} \\ \textbf{(W/kg)}} & \textbf{PPR} & 
        \shortstack{\textbf{$\tau_{MAX}$} \\ \textbf{(Nm/kg)}} &
        \shortstack{\textbf{$\tau_{RMS}$} \\ \textbf{(Nm/kg)}} \\ \hline
        
    \multirow{2}{*}{Hip} 
    & 25 & 0.267 & -0.052 & 0.837 & 0.372 & 0.246 \\ \cline{2-7}
    & 50 & 0.359 & -0.079 & 0.820 & 0.729 & 0.394 \\ \hline
    
    \multirow{2}{*}{Knee} 
    & 25 & 0.163 & -0.084 & 0.660 & 0.310 & 0.162 \\ \cline{2-7}
    & 50 & 0.168 & -0.082 & 0.672 & 0.339 & 0.167 \\ \hline
    
    \multirow{2}{*}{Ankle} 
    & 25 & 0.124 & -0.032 & 0.795 & 0.372 & 0.181 \\ \cline{2-7}
    & 50 & 0.205 & -0.033 & 0.861 & 0.745 & 0.283 \\ \hline
    
    \end{tabular}
\end{table}

%% file: appendix.tex
\appendix

\section{Appendix}\label{sec:appendix}

\input{max_iso}

\input{generic_muscle_prop}

\input{COT_Reduction_PerSubject}

\input{stat}


\begin{figure}[htb!]
    \centering
    \includegraphics[width=0.90\textwidth]{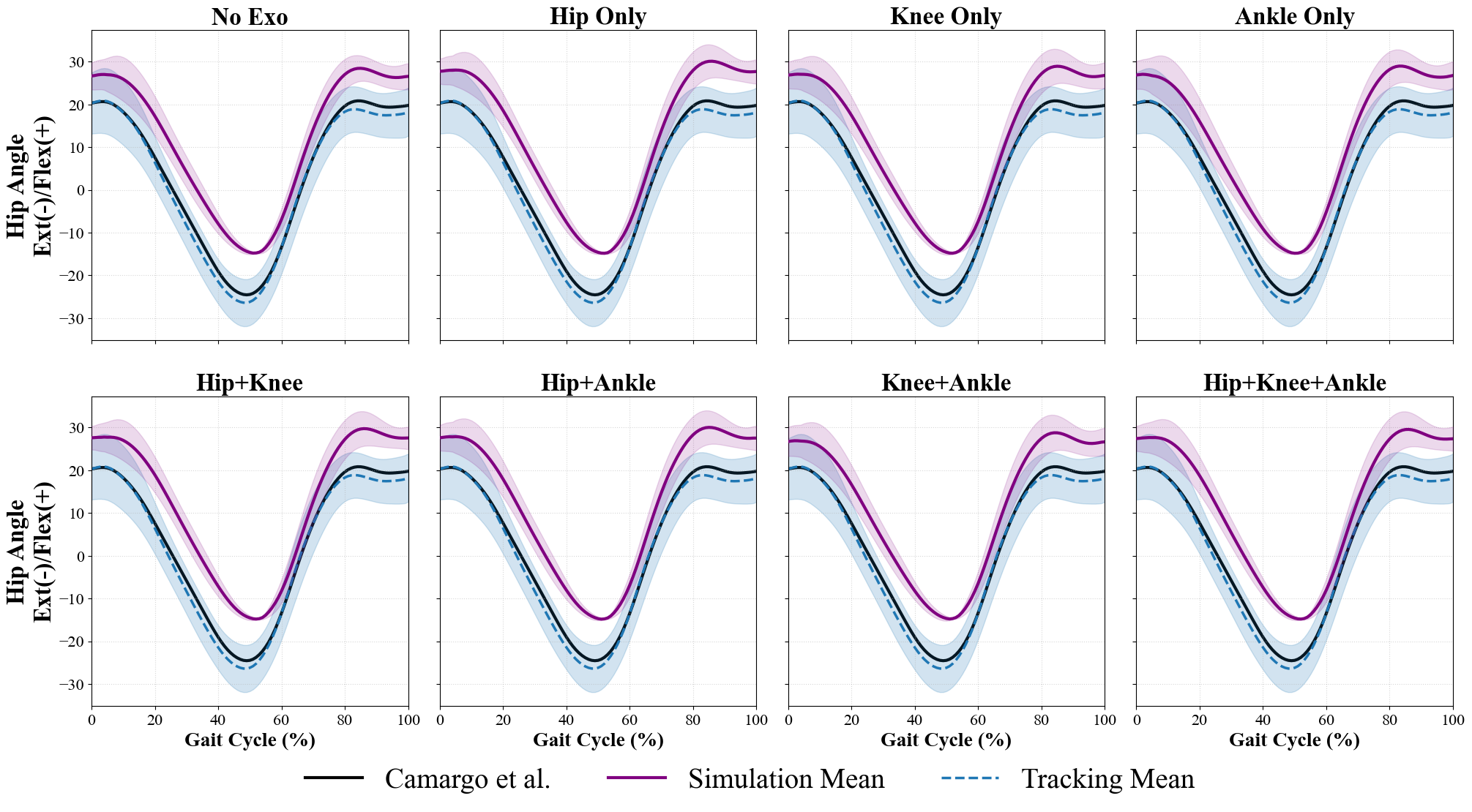}
    \caption{Hip flexion-extension angles resulting from predictive tracking simulations across seven exoskeleton assistance conditions (25 Nm) and the unassisted baseline (NoExo). Each subplot shows the hip angle profile obtained by solving the optimal control problem for the specific condition.}
    \label{fig:hip_angle_25Nm}
\end{figure}

\begin{figure}[htb!]
    \centering
    \includegraphics[width=0.90\textwidth]{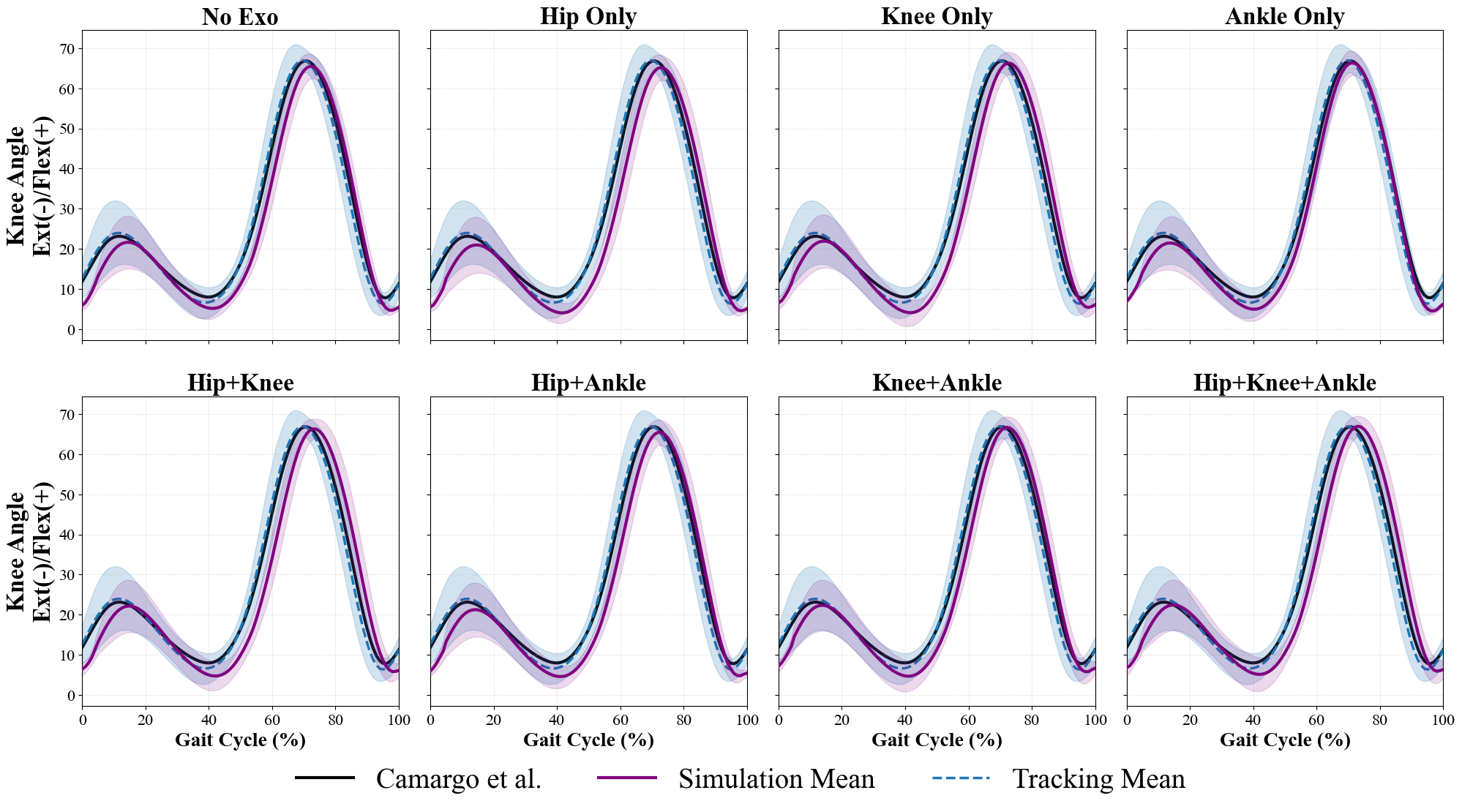}
    \caption{Knee flexion-extension angles resulting from predictive tracking simulations across seven exoskeleton assistance conditions (25 Nm) and the unassisted baseline (NoExo). Each subplot shows the knee angle profile obtained by solving the optimal control problem for the specific condition.}
    \label{fig:knee_angle_25Nm}
\end{figure}

\begin{figure}[htb!]
    \centering
    \includegraphics[width=0.90\textwidth]{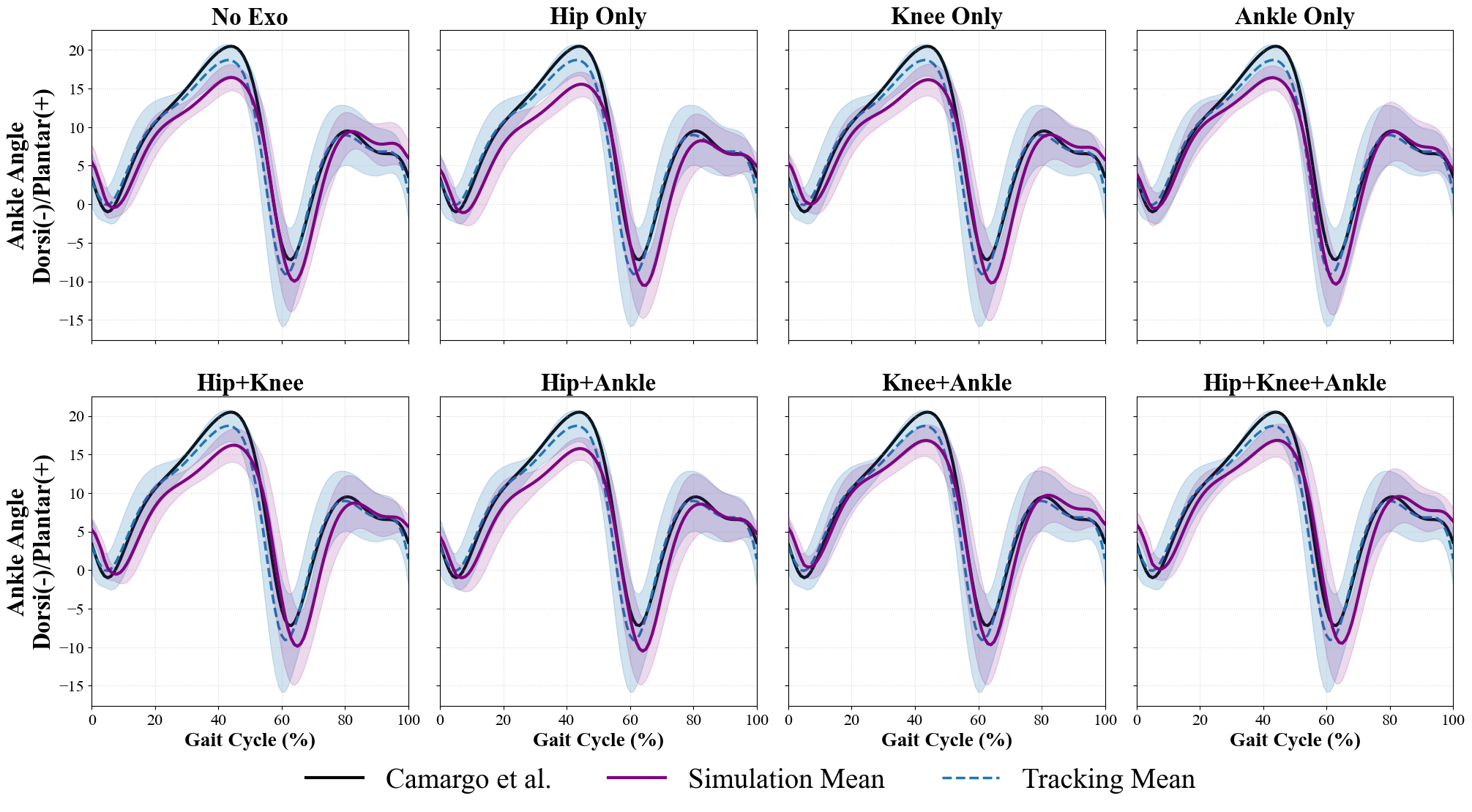}
    \caption{Ankle dorsiflexion-plantarflexion angles resulting from predictive tracking simulations across seven exoskeleton assistance conditions (25 Nm) and the unassisted baseline (NoExo). Each subplot shows the ankle angle profile obtained by solving the optimal control problem for the specific condition.}
    \label{fig:ankle_angle_25Nm}
\end{figure}

\begin{figure}[htb!]
    \centering
    \includegraphics[width=0.80\textwidth]{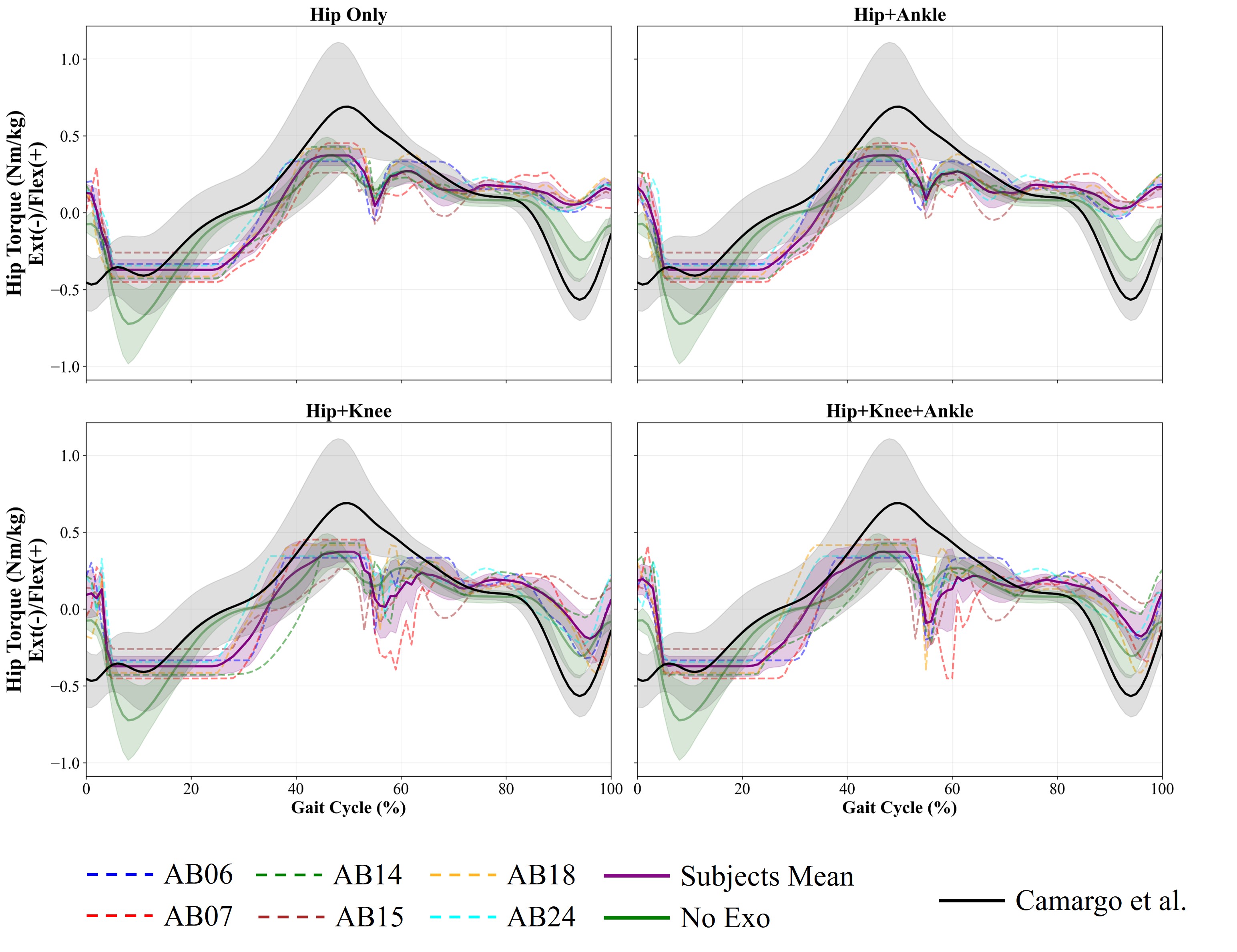}
    \caption{Predicted hip assistive torque (25 Nm) profiles across the gait cycle for the four exoskeleton configurations (Hip Only, H+A, H+K, and H+K+A) that incorporate the hip.}
    \label{fig:hip_torque_25Nm}
\end{figure}

\begin{figure}[htb!]
    \centering
    \includegraphics[width=0.80\textwidth]{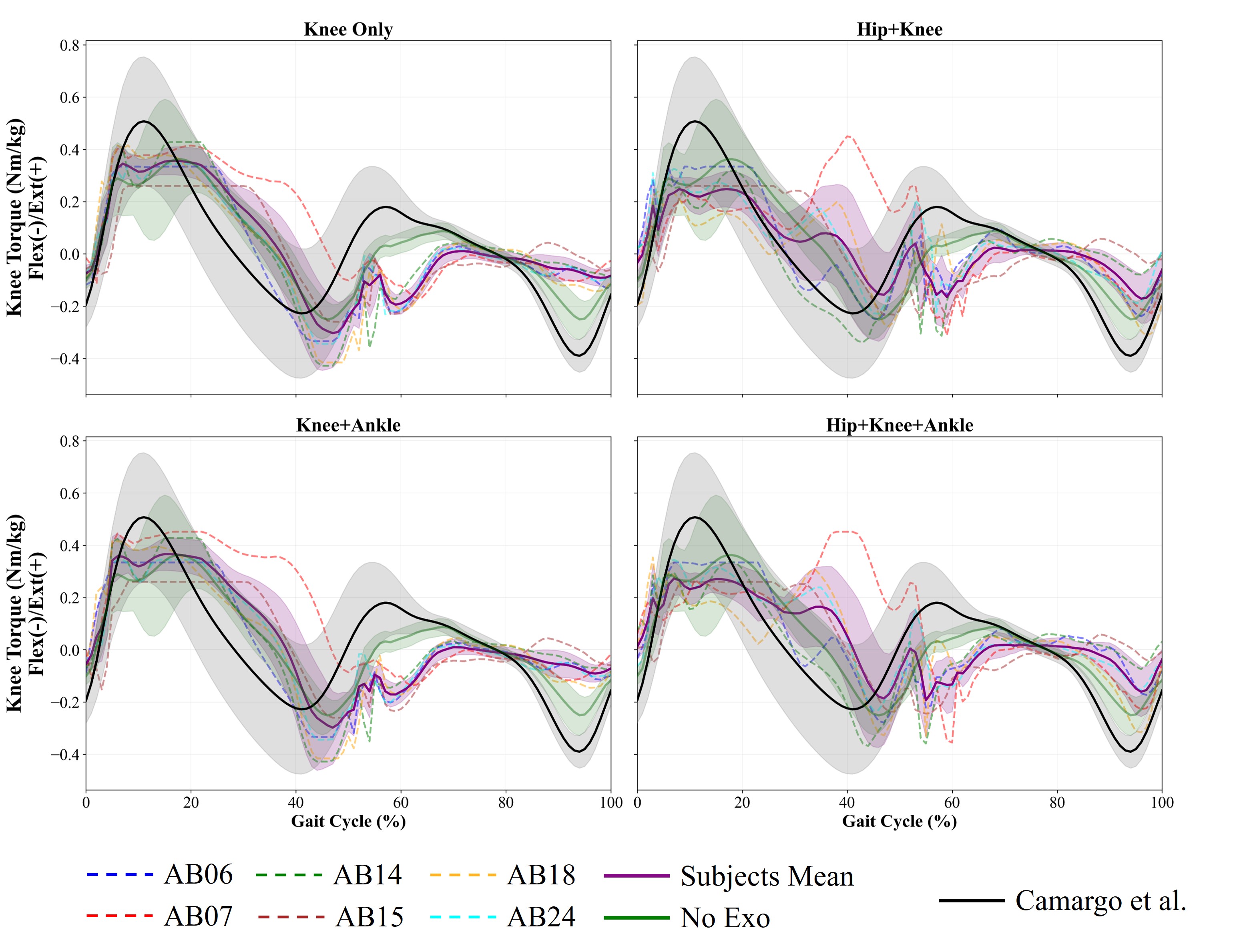}
    \caption{Predicted knee assistive torque (25 Nm) profiles across the gait cycle for the four exoskeleton configurations (Knee Only, H+K, K+A, and H+K+A) that incorporate the knee.}
    \label{fig:knee_torque_25Nm}
\end{figure}

\begin{figure}[htb!]
    \centering
    \includegraphics[width=0.80\textwidth]{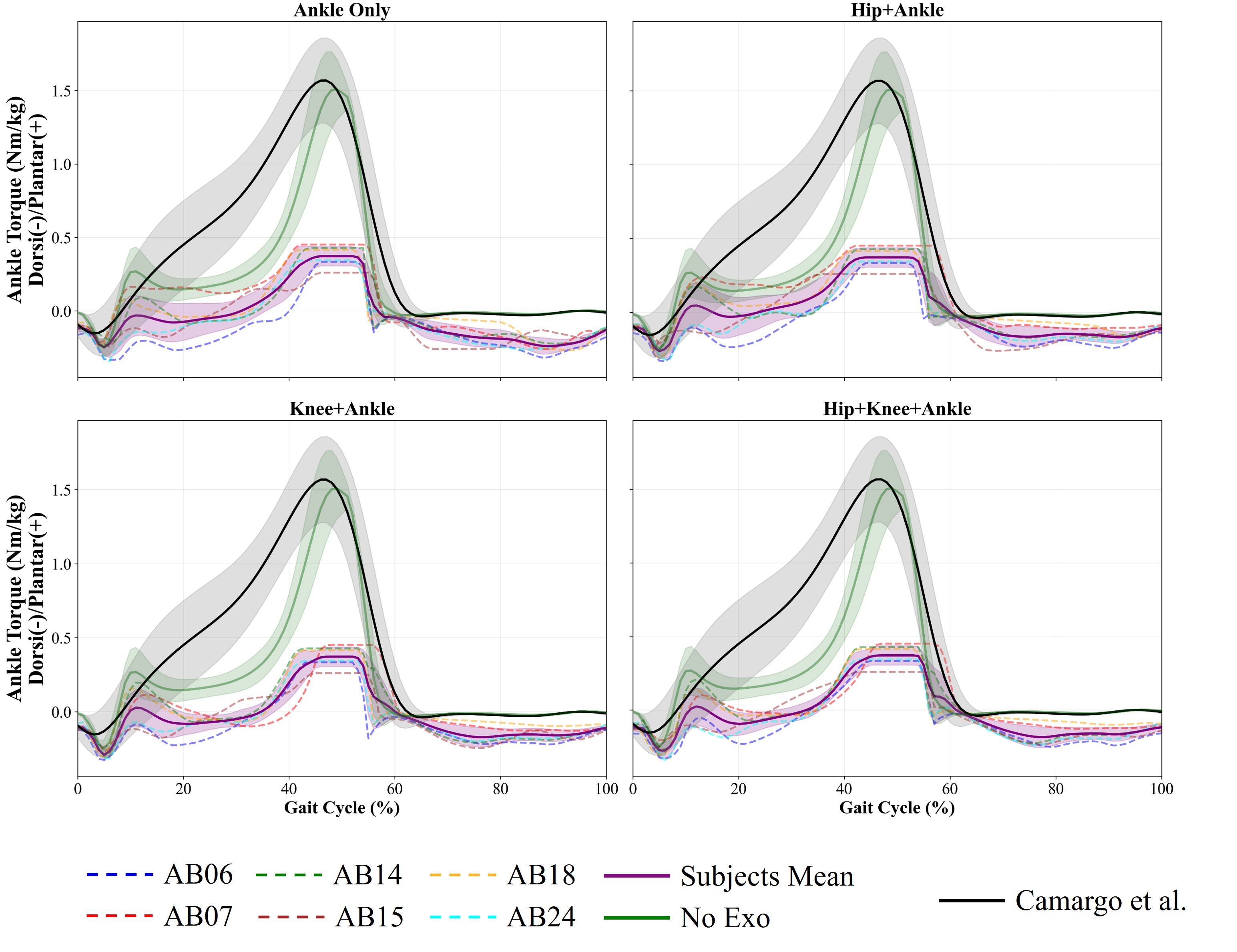}
    \caption{Predicted ankle assistive torque (25 Nm) profiles across the gait cycle for the four exoskeleton configurations (Ankle Only, H+A, K+A, and H+K+A) that incorporate the ankle.}
    \label{fig:ankle_torque_25Nm}
\end{figure}

\begin{figure}[htb!]
    \centering
    \includegraphics[width=0.80\textwidth]{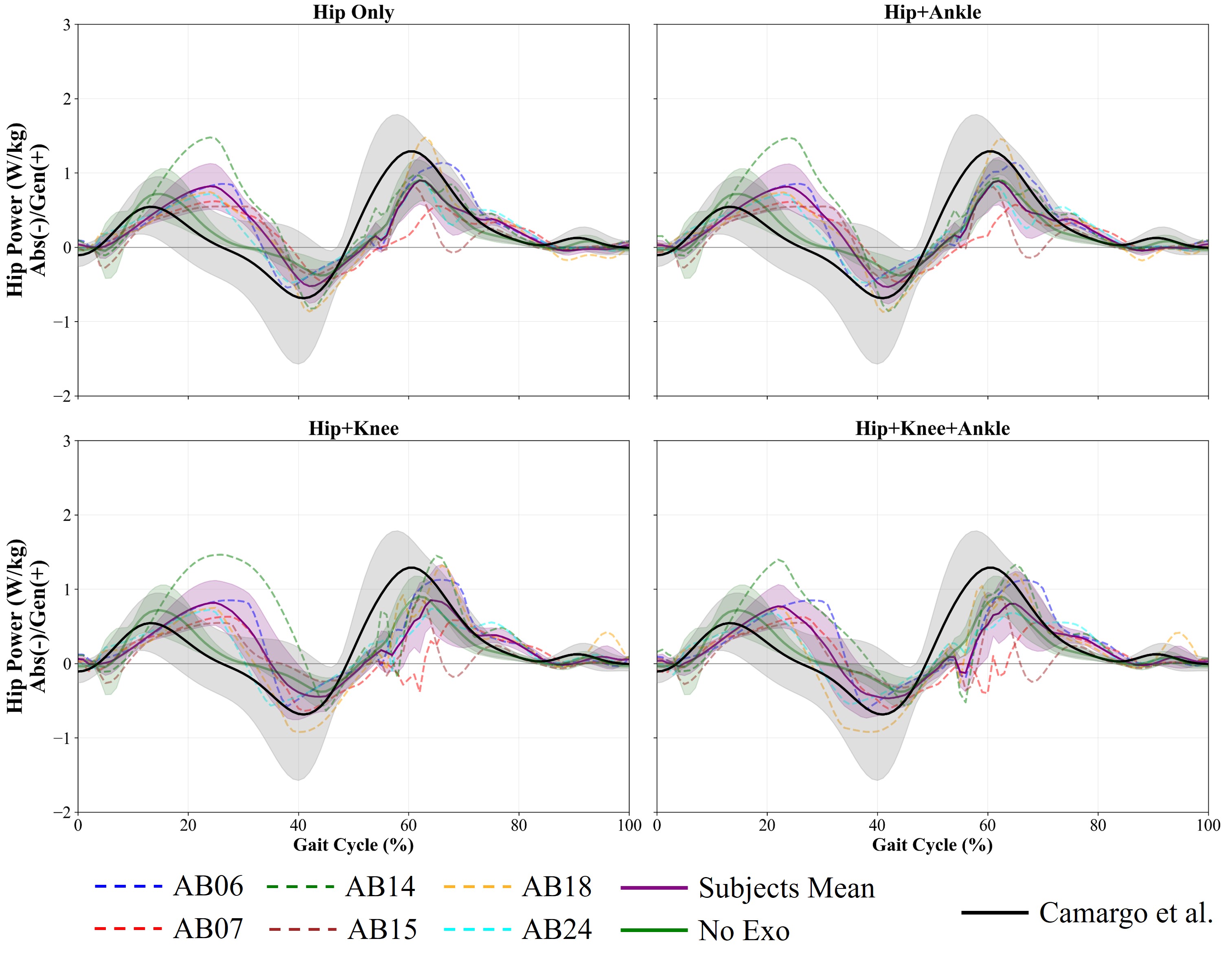}
    \caption{Predicted hip joint power under 25 Nm assistance across the gait cycle for the four exoskeleton configurations (Hip Only, H+A, H+K, and H+K+A) that incorporate the hip.}
    \label{fig:hip_power_25Nm}
\end{figure}

\begin{figure}[htb!]
    \centering
    \includegraphics[width=0.80\textwidth]{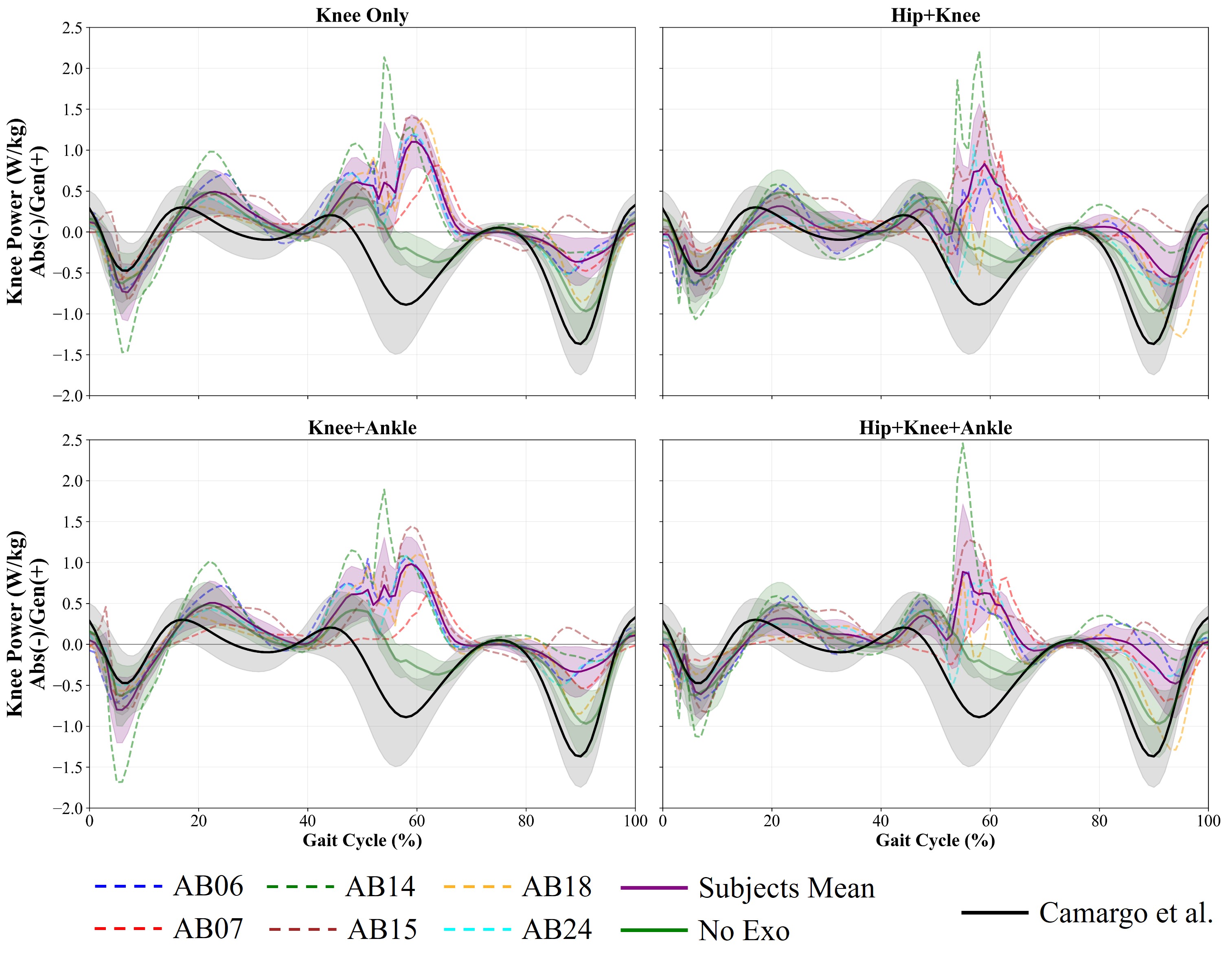}
    \caption{Predicted knee joint power under 25 Nm assistance across the gait cycle for the four exoskeleton configurations (Knee Only, H+K, K+A, and H+K+A) that incorporate the knee.}
    \label{fig:knee_power_25Nm}
\end{figure}

\begin{figure}[htb!]
    \centering
    \includegraphics[width=0.80\textwidth]{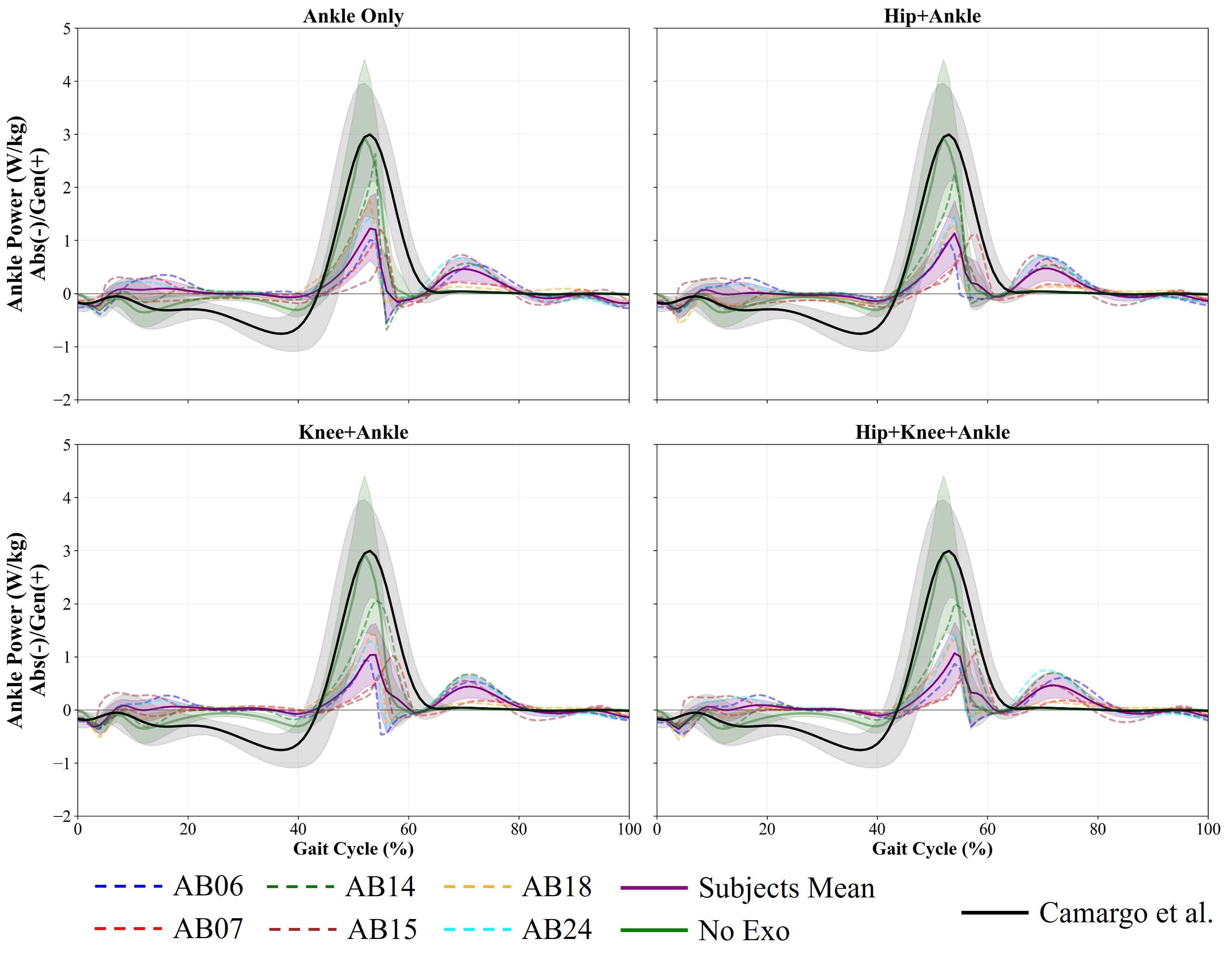}
    \caption{Predicted ankle joint power under 25 Nm assistance across the gait cycle for the four exoskeleton configurations (Ankle Only, H+A, K+A, and H+K+A) that incorporate the ankle.}
    \label{fig:ankle_power_25Nm}
\end{figure}

    \begin{figure}[htb!]
    \centering
    \includegraphics[width=0.99\textwidth]{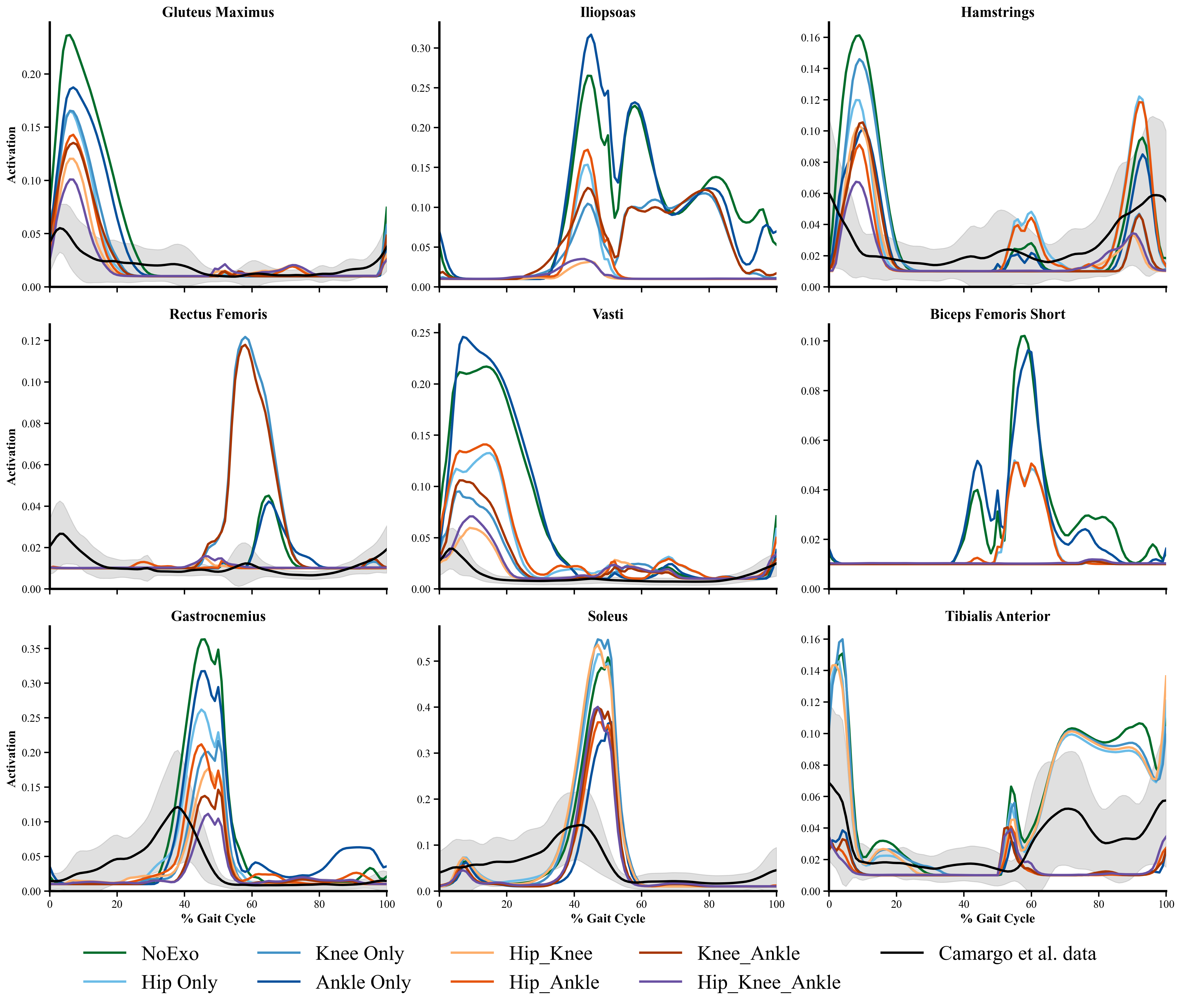}
    \caption{Muscle activation profiles across nine lower-limb muscles over the gait cycle, averaged across subjects, under NoExo and seven exoskeleton assistance conditions (Hip Only, Knee Only, Ankle Only, H+K, H+A, K+A, H+K+A) at $25~\mathrm{Nm}$ assistive torque. 
    *Note: The experimental data for iliopsoas and biceps femoris short head was not available.}
    \label{fig:emg_25Nm}
\end{figure}

%% file: max_iso.tex
\begin{table}[htb!]
\centering
\caption{Maximum isometric muscle forces (N) for the subject-specific models obtained by scaling the generic model values based on subject anthropometry.}
\label{tab:max_isometric_muscle_forces}
\begin{tabular}{lcccccc}
\toprule
Muscle & AB06 & AB07 & AB14 & AB15 & AB18 & AB24 \\
\midrule
HAMS & 3071 & 2042 & 2990 & 5421 & 1846 & 3243 \\
BFSH &  914 &  608 &  890 & 1614 &  550 &  966 \\
GMAX & 2211 & 1470 & 2152 & 3903 & 1329 & 2335 \\
IL   & 2664 & 1771 & 2593 & 4702 & 1601 & 2813 \\
RF   & 1330 &  884 & 1294 & 2347 &  799 & 1404 \\
VAS  & 5687 & 3781 & 5536 & 10038 & 3419 & 6006 \\
GAS  & 2843 & 1891 & 2768 & 5019 & 1709 & 3003 \\
SOL  & 5842 & 3885 & 5688 & 10313 & 3513 & 6170 \\
TA   & 3412 & 2269 & 3322 & 6023 & 2051 & 3603 \\
\bottomrule
\end{tabular}
\end{table}

%% file: generic_muscle_prop.tex
\begin{table}[htb!]
\centering
\caption{Key muscle properties for the generic model.}
\label{tab:generic_model_muscle_properties}
\begin{tabular}{lcccc}
\toprule
Muscle & Max Iso. Force (N) & Opt. Fiber (m) & Slack Length (m) & Pennation (rad) \\
\midrule
HAMS & 2700 & 0.1091 & 0.3262 & 0.0000 \\
BFSH & 804  & 0.1719 & 0.0884 & 0.4014 \\
GMAX & 1944 & 0.1548 & 0.0723 & 0.0000 \\
IL   & 2342 & 0.0975 & 0.1560 & 0.1396 \\
RF   & 1169 & 0.1142 & 0.3104 & 0.0873 \\
VAS  & 5000 & 0.1077 & 0.1168 & 0.0524 \\
GAS  & 2500 & 0.0869 & 0.3477 & 0.2967 \\
SOL  & 5137 & 0.0482 & 0.2409 & 0.4363 \\
TA   & 3000 & 0.0938 & 0.2134 & 0.0873 \\
\bottomrule
\end{tabular}
\end{table}

%% file: COT_Reduction_PerSubject.tex
\begin{table}[h]
\centering
\footnotesize
\caption{Percentage reduction in COT relative to the unassisted (NoExo) condition for each subject across assistance configurations at 25, and 50 Nm}
\label{tab:COT_subjectwise}
\begin{tabular}{|l|c|c|c|c|c|c|}
\hline
\textbf{Condition} & \textbf{AB06} & \textbf{AB07} & \textbf{AB14} & \textbf{AB15} & \textbf{AB18} & \textbf{AB24} \\ \hline

\multicolumn{7}{|c|}{\textbf{25 Nm}} \\ \hline
Hip Only  & 25.85 & 27.51 & 25.72 & 15.82 & 30.10 & 21.62 \\
Knee Only & 16.08 & 17.10 & 18.13 & 16.61 & 20.10 & 15.18 \\
Ankle Only & 12.16 & 11.19 & 14.41 & 8.64  & 11.51 & 13.21 \\
H+K   & 32.88 & 32.34 & 33.42 & 26.16 & 34.73 & 26.31 \\
H+A   & 36.25 & 35.81 & 37.35 & 23.50 & 38.84 & 31.94 \\
K+A   & 27.23 & 29.46 & 30.96 & 24.67 & 32.43 & 27.78 \\
H+K+A & 42.96 & 42.65 & 43.80 & 34.16 & 45.54 & 38.10 \\ \hline

\multicolumn{7}{|c|}{\textbf{50 Nm}} \\ \hline
Hip   & 35.44 & 30.31 & 34.80 & 23.22 & 37.44 & 29.42 \\
Knee  & 19.38 & 17.80 & 19.36 & 20.57 & 21.86 & 16.03 \\
Ankle & 15.83 & 17.00 & 21.75 & 10.84 & 17.67 & 19.05 \\
H+K   & 37.78 & 34.64 & 37.31 & 32.38 & 38.37 & 30.42 \\
H+A   & 48.20 & 41.34 & 51.55 & 32.94 & 50.86 & 42.72 \\
K+A   & 35.88 & 34.62 & 41.26 & 31.67 & 40.02 & 34.49 \\
H+K+A & 51.27 & 46.38 & 54.65 & 42.22 & 50.85 & 45.63 \\ \hline

\end{tabular}
\end{table}

%% file: stat.tex
\begin{table}[htbp]
\centering
\caption{Statistical comparisons of COT reduction among assistance
configurations at $50~\mathrm{Nm}$.}
\label{tab:cot_minimal_statistics}

\begin{tabular}{p{0.27\linewidth}p{0.24\linewidth}ccc}
\toprule
Comparison & Test & Statistic & Raw $p$ & Holm-adjusted $p$ \\
\midrule

H, K, A
& Exact Friedman
& $Q=9.33$
& 0.0055
& \textbf{0.0166 (significant)} \\

H+K, H+A, K+A
& Exact Friedman
& $Q=9.00$
& 0.0081
& \textbf{0.0166 (significant)} \\

H+K+A versus H+A
& Exact paired Wilcoxon signed-rank
& $W=1.00$
& 0.0625
& 0.0625 (not significant) \\

\bottomrule
\end{tabular}

\begin{minipage}{0.94\linewidth}
\footnotesize
\textit{Note.} Analyses used paired observations from six
subject-specific models. Exact $p$-values were calculated, and Holm
correction was applied across the three tests. Bold values indicate
Holm-adjusted $p<0.05$.
\end{minipage}
\end{table}